%% file: main.tex
\documentclass[runningheads]{llncs}
\usepackage{bbding}
\input{macros}

\begin{document}

\title{Synthesizing Efficiently Monitorable Formulas in Metric Temporal Logic}%\thanks{The first two authors contributed equally, the remaining authors are ordered alphabetically}}%

\titlerunning{Synthesizing Efficiently Monitorable Formulas in MTL}

%\author{Anonymous Authors}
%\authorrunning{}
%\institute{}

 \author{Ritam Raha\inst{1,2}\and Rajarshi Roy\inst{3}, Nathana\"el Fijalkow\inst{2} \and Daniel Neider \inst{4,5}\and Guillermo A. P\'erez\inst{1}
 } 

\authorrunning{Raha et al.}

\institute{University of Antwerp, Flanders Make, Belgium\\
%\email{\{guillermoalberto.perez,ritam.raha\}@uantwerpen.be}
\and LaBRI, University of Bordeaux, Bordeaux, France\\
%\email{nathanael.fijalkow@gmail.com} 
\and
Max Planck Institute for Software Systems, Kaiserslautern, Germany\\
%\email{rajarshi@mpi-sws.org} 
\and  Technical University of Dortmund, Germany\\
\and Center for Data Science and Security, University Alliance Ruhr, Germany
%\email{daniel.neider@cs.tu-dortmund.de}
} 
\maketitle

\begin{abstract}
In runtime verification, manually formalizing a specification for monitoring system executions is a tedious and error-prone process. To address this issue, we consider the problem of automatically synthesizing formal specifications from system executions. To demonstrate our approach, we consider the popular specification language Metric Temporal Logic ($\MTL$) which is particularly tailored towards specifying temporal properties for cyber-physical systems (CPS). 
Most of the classical approaches for synthesizing temporal logic formulas aim at minimizing the size of the formula.
However, for efficiency in monitoring, along with the size, the amount of ``lookahead" required for the specification becomes relevant, especially for safety-critical applications. 
We formalize this notion and devise a learning algorithm that synthesizes concise formulas having bounded lookahead. 
To do so, our algorithm reduces the synthesis task to a series of satisfiability problems in Linear Real Arithmetic (LRA) and generates $\MTL$ formulas from their satisfying assignments. 
The reduction uses a novel encoding of a popular $\MTL$ monitoring procedure using LRA. 
%We also present complexity results for the corresponding decision problems. 
Finally, we implement our algorithm in a tool called~\tool{} and demonstrate its ability to synthesize efficiently monitorable $\MTL$ formulas in a CPS application.
\end{abstract}

%\keywords{Metric Temporal Logic, Runtime Verification, Specification Mining, Interpretability, Formal Methods}

%%%%%%%%%%%%%%%%%%%%%%%%%%%%%%%%%%%%%%%%%%%%%%%%%%%%%%%%%%%%%%%%%%%%%%%%

\section{Introduction}

\input{intro}

\input{prelims}
%%%%%%%%%%%%%%%%%%%%%%%%%%%%%%%%%%%%%%%%%%%%%%%%%%%%%%%%%%%%%%%%%%%%%%%%

\input{algorithms}
\section{Experiments} 
%%%%%%%%%%%%%%%%%%%%%%%%%%%%%%%%%%%%%%%%%%%%%%%%%%%%%%%%%%%%%%%%%%%%%%%%
\input{experiments}

%%% The acknowledgments section is defined using the "acks" environment
%%% (rather than an unnumbered section). The use of this environment 
%%% ensures the proper identification of the section in the article 
%%% metadata as well as the consistent spelling of the heading.

\section{Discussion and Conclusion}
\input{conclusion}
%%%%%%%%%%%%%%%%%%%%%%%%%%%%%%%%%%%%%%%%%%%%%%%%%%%%%%%%%%%%%%%%%%%%%%%%

%\section{Acknowledgement}
%We acknowledge...
%%%%%%%%%%%%%%%%%%%%%%%%%%%%%%%%%%%%%%%%%%%%%%%%%%%%%%%%%%%%%%%%%%%%%%%%

%%% The next two lines define, first, the bibliography style to be 
%%% applied, and, second, the bibliography file to be used.
\newpage
\bibliography{refs}
%%%%%%%%%%%%%%%%%%%%%%%%%%%%%%%%%%%%%%%%%%%%%%%%%%%%%%%%%%%%%%%%%%%%%%%%

\newpage
\appendix
\input{appendix}
\end{document}

%% file: macros.tex
\usepackage{complexity}
\usepackage{framed}
\usepackage[english]{babel}
\usepackage{amsmath,amssymb,amsfonts,mathtools}
\usepackage{color}
\usepackage{graphicx}
\usepackage{subcaption}
\usepackage{url}
\usepackage[hidelinks]{hyperref}
\usepackage{authblk}
\usepackage{relsize}
\usepackage{graphicx}
\usepackage{algorithm}
\usepackage[noend]{algpseudocode}
\usepackage{tikz}
\usepackage{pgfplots}
\usepackage{multirow}
\usepackage{ntheorem}
\usepackage{stackengine,scalerel}
\usepackage[strings]{underscore}
\usepackage{paralist}
\usepackage{booktabs}

\usepackage{wrapfig}
\newcommand{\fr}{\mathit{fr}}
\usepackage{mathtools}
\usepackage{scalerel}
\newcommand{\shortiff}{\mathrel{\scalerel*{\iff}{\scriptstyle\Rightarrow\Leftarrow}}}
\newcommand{\fmodels}{\models_{\mathrm{f}}}

\newcommand{\N}{\mathbb{N}}

\renewcommand{\R}{\mathbb{R}}

\renewcommand{\vec}[1]{\boldsymbol{#1}}

\usepackage[]{todonotes}

\newcommand{\LTL}{\textrm{LTL}}
\newcommand{\STL}{\textrm{STL}}
\newcommand{\MTL}{\textrm{MTL}}

\newcommand{\prop}{\mathcal{P}}
\DeclareMathOperator{\lF}{\mathbf{F}}
\DeclareMathOperator{\lG}{\mathbf{G}}
\DeclareMathOperator{\lU}{\mathbf{U}}

\DeclareMathOperator{\false}{\mathit{false}}
\DeclareMathOperator{\true}{\mathit{true}}

\newcommand{\dom}{\mathbb{T}}
\newcommand{\gsep}{\mathbf{G}\text{-}\mathrm{sep}}
\newcommand{\gksepable}{K\text{-}\mathit{infix}\text{-}\mathit{separable}}
\newcommand{\finaltime}{T}
\newcommand{\sample}{\mathcal{S}}

\newcommand{\bound}{K}

\newcommand{\prefix}[1]{\vec{#1}_\dom}
\newcommand{\finalsize}{n}

\newcommand{\syn}{\textsc{SynTL}}
\newcommand{\synd}{\textsc{SynTL}_{d}}

\newcommand{\I}{\mathcal{I}}

\newcommand{\inter}{\iota}
\newcommand{\fullcons}{\Phi^{n}_{\sample,\bound}}
\newcommand{\strcons}{\Phi^{\mathit{str}}}
\newcommand{\frcons}{\Phi^{\fr}}
\newcommand{\semcons}{\Phi^{\mathit{sem}}}
\newcommand{\ivalset}[2]{\mathcal{I}^{#1}_{#2}({\prefix{x}})}
\newcommand{\ivalsetx}[2]{\mathcal{I}^{#1}_{#2}({\prefix{x}^s})}
\newcommand{\ivalsetu}[2]{\mathcal{I}^{#1}_{#2}(u_1)}

\newcommand{\tool}{\texttt{TEAL}}

%% file: intro.tex
% Outline
% 1. Runtime verification, software & hardware
% 2. Two big problems: specifications hard to get, specifications might not be
%    easy to monitor
% 3. Learning specifications: lots of work in LTL, but signals and time are
%    useful (hence how active research on $\STL$ is), we focus on $\MTL$
% 4. Quantifying how easy a formula can be monitored: future reach
% 5. Our problem: Efficiently monitorable $\MTL$ formulas (define sample,
%    important - prefixes!)
% 6. Related work and contributions
\label{sec:intro}
Runtime verification is a well-established method for ensuring the correctness of cyber-physical systems during runtime.
Techniques in runtime verification are known to be more rigorous than conventional testing while not being as resource-intensive as exhaustive formal verification~\cite{rv-2022}.
In the field of runtime verification, among other techniques, monitoring system executions against formal specifications during runtime is a widely used one.
Over the years, numerous monitoring techniques have been proposed for a variety of specification languages~\cite{HavelundP18,DonzeFM13,DeshmukhDGJJS15,BartocciDDFMNS18}.

In this work, we focus on Metric Temporal Logic ($\MTL$)~\cite{Koymans90}---a specification language popularly employed for monitoring cyber-physical systems~\cite{HoOW14,maler}.
$\MTL$ is a real-time extension of Linear Temporal Logic (LTL)~\cite{Pnueli77} augmented with timing constraints for temporal operators.
$\MTL$ specifications are often easy to interpret due to their resemblance to natural language and, thus, also find applications in Artificial Intelligence~\cite{ZheMTL}.
While there are many possible semantics of $\MTL$ (e.g., discrete, dense-time pointwise, etc.~\cite{JoelMTL}), we employ the dense-time continuous semantics as it is more natural and general than the counterparts~\cite{BasinKZ11,BaldorN12}.
We expand on $\MTL$ and other prerequisites in Section~\ref{subsec:prelims}.

Virtually all verification techniques for $\MTL$ rely on the availability of a formal specification.
However, manually writing specifications is a tedious and error-prone task~\cite{AmmonsBL02}.
Synthesizing functional, correct, and interpretable specifications that precisely express the design requirements has been one of the major challenges in the adoption of formal techniques for verification~\cite{BjornerH14,Rozier16}.

%We posit that runtime verification techniques for cyber-physical systems --- that is, systems in which a physical mechanism interacts with computer-based
%algorithms --- currently have two shortcomings. First, . However, manually formalizing a specification that
%expresses precisely the design requirements is notoriously difficult and error-prone . Even worse, the training effort required to reach
%proficiency with specification languages can be disproportionate to the
%expected benefits. Second, and also related to the specification
%effort, different correctness specifications (in any specification formalism)
%may impose different efficiency constraints on monitors for them. 
%To be more
%concrete, a concise specification is desirable as it makes the process easier and also they become human-interpretable. 
To tackle the lack of formal specifications, there have been efforts to automatically synthesize
specifications from system executions.
Most of the existing works have targeted specification languages such as Linear Temporal Logic (LTL)~\cite{camachomcilraith2021,flie,scarlet} and Signal Temporal Logic (STL)~\cite{asarin,linardJana,MohammadinejadD20,abs-1711-06202}, with few works for MTL~\cite{HoxhaDF18,ZheMTL}.
Many of the works tend to synthesize specifications that are \emph{concise} in size.
Concise specifications are preferred over large ones because, based on the principle of Occam's razor, they are easier for humans to understand~\cite{0002FN20}.
%
%%Intuitively, the
%problem asks whether there exists a specification such that all elements from a given set of positive examples satisfy it and all those from a given set of negative examples do not satisfy it. Additionally, in the spirit of Occam's
%razor, it has been argued that consistent specifications that are
%\emph{concise} --- i.e., they have a smaller parse tree --- are preferred over
%larger more complex possibilities. 
%

However, conciseness is not the only measure of interest for specifications, especially in the context of online monitoring.
In online monitoring, specifically in \emph{stream-based} runtime monitoring, a monitor reads an execution as a stream of data and verifies if a given specification is invariant (i.e., holds at all time points) in the execution.
Many stream-based monitors~\cite{GorostiagaS21,KaneCDK15,LimaHRTY23} support $\MTL$ formulas.
Typically, such monitors produce a stream of (Boolean) verdicts with some ``latency'', which depends on the lookahead of the formula.
The lookahead required for an $\MTL$ formula is often formalized as its \emph{future-reach}~\cite{HoOW14,hunter}, which is the amount of time required to determine its satisfaction at any time point.

%\todo{Guillermo, could you help in writing here? Add that we are interested in stream-based monitoring} 
%In online monitoring, one looks at a stream of events from a system during its execution and examines whether the specification remains invariant during all time points in the execution. 
%To reduce latency, a specification must not only be concise but also require a small amount of ``lookahead'' to determine if an execution is good or bad.
%The amount of lookahead required is typically formalized as the \emph{future-reach}~\cite{HoOW14,hunter} of a formula, which is the minimal time required to determine the satisfaction of the formula at any given time.

%One of the specification languages that is popularly employed in cyber-physical systems for specifying temporal properties is temporal logic. 
%They are typically preferred as it is very close to natural language. This is why, temporal logic has been used as the specification language in a number of monitoring tools~\cite{Drusinsky00,Havelund2001JavaPA}. In this work, we 
%focus on Metric Temporal logic, $\MTL$ for short, which can be used for describing dense-time properties of reactive systems.
With the aim of reducing the latency for efficient online monitoring, we focus on automatically synthesizing $\MTL$ specifications based on two regularizers, size and future-reach. 
As input data, we rely on a sample $\sample$ consisting of executions of a system that are observed for a finite duration.
We consider the sample to be partitioned into a set $P$ of positive (or desirable) executions and a set $N$ of negative (or undesirable) executions.

We now formulate the central problem of synthesizing MTL formulas as follows:
given a sample $\sample = (P,N)$ and a future-reach bound $\bound$, synthesize a minimal size $\MTL$ formula $\varphi$ that (i) is \emph{globally-separating} for $\sample$, in that $\varphi$ holds at all time points in the positive executions and does not hold at some time point in the negative executions, and (ii) the future-reach of $\varphi$ is smaller than $\bound$.
The property of being globally-separating for $\sample$ ensures that prospective formula $\varphi$ is invariant in the desirable executions and not in the undesirable executions, as is typically preferred in specifications for online monitoring~\cite{aerialtool}.
We expand on the problem formulation in Section~\ref{subsec:problem-form}.

Also, interestingly, without a future-reach bound, the most concise $\MTL$ formula that can be synthesized can have a large future-reach value, increasing the latency required for online monitoring. To illustrate this, assume that we observe some simulations of an autonomous vehicle.
During the simulations, we sample executions (shown below) of the vehicle every second for six seconds.
We classify them as positive (denoted using  $u_i$'s) or negative (denoted using $v_i$'s) based on whether the vehicle encountered a collision or not.
\begin{center}
\begin{tabular}{ l c c c c c c }
& 0 & 1 & 2 & 3 & 4 & 5 \\
$u_1$: & $\{p,q\}$ & $\{p\}$ & $\{q\}$ & $\{p,q\}$ & $\{p\}$ & $\{p\}$\\
$u_2$: & $\{q\}$ & $\{\}$ & $\{q\}$ & $\{p\}$ & $\{p\}$ & $\{p,q\}$\\
$v_1$: & $\{p\}$ & $\{q\}$ & $\{\}$ & $\{\}$ & $\{\}$ & $\{\}$\\
$v_2$: & $\{p\}$ & $\{p,q\}$ & $\{p\}$ & $\{\}$ & $\{p\}$ & $\{\}$\\
\end{tabular}
\end{center}
In the executions, we use $p$ to denote that there is no obstacle within a particular unsafe distance ahead of the vehicle and $q$ to denote that the vehicle's brake is triggered. 
Our setting considers executions to be \emph{continuous}. Thus, to ensure continuity of execution, in the above example, if $p$ occurs at time point $t$, we interpret it as $p$ holding during the entire interval $[t,t+1)$. 
We also assume that the executions last up to a final time point $T$ which is 6 for this example.
Thus, for the execution $u_1$, $p$ holds in the intervals $[0,2)$ and $[3,6)$.

In the sample, a minimal globally separating formula is $\varphi_1 = \lF_{[0,3]} q$.
The formula $\varphi_1$ being globally separating indicates that in all positive executions, 
the brake is triggered every three seconds (i.e., within the interval $[t,t+3]$ for every time point $t$), irrespective of whether there is an obstacle within the unsafe distance.
The formula $\varphi_1$ has size two and a future-reach of three seconds, meaning that any online monitor requires a three second lookahead window to check the satisfaction of $\varphi_1$.
There is another formula $\varphi_2 = \neg p \rightarrow \lF_{[0,1]} q$ that is globally separating for the sample.
The formula $\varphi_2$ being globally-separating indicates that in all positive executions, for every time point $t$, if an obstacle is within the unsafe distance, then the brake is triggered within one second (i.e., within the interval $[t,t+1]$).
Although of size five, $\varphi_2$ has future-reach of one second and will be typically preferred over $\varphi_1$ for online monitoring in a safety-critical scenario.

%This example motivates our choice of incorporating the future-reach into our learning process.
%Importantly, we have chosen to focus on reactive
%cyber-physical systems and thus model their executions as infinite %This means that the ``sample'', in our problem described above, consists of
%prefixes of infinite executions (importantly, we do not consider standard finite-word
%semantics for $\MTL$). We will discuss this point in more detail later when we introduce the semantics.

For the problem of synthesizing MTL formulas, we first study whether a solution exists.
It turns out that there are samples $\sample$ and future-reach bound $K$ for which there might not exist any formula that is globally-separating for $\sample$ and has future-reach within $K$.
To aid in checking whether a prospective formula exists, we identify a simple characterization of $\sample$ based on the future-reach $K$.
Such a characterization enables us to design an $\NP$ algorithm that can decide whether a prospective algorithm exists.
Also, it provides an upper-bound, which is polynomial in the inputs $\sample$ and $K$, on the size of the prospective formula if one exists.
We mention the details of the existence check in Section~\ref{sec:existence}.

To synthesize a prospective formula, we rely on a reduction to constraint satisfaction problems. 
In particular, following other works in synthesis of formulas~\cite{flie,0002FN20}, our algorithm encodes the problem in a series of satisfiability modulo theory (SMT) problems in Linear Real Arithmetic (LRA).
To our knowledge, we design the first SMT-based algorithm that can synthesize MTL formulas of arbitrary syntactic structure.
Such an SMT-based algorithm allows us to extend our algorithm to work for other settings that are common in the synthesis of formulas~\cite{GaglioneNRTX21,abs-2206-06722}.

Further, we analyze the complexity of the decision version of the problem of synthesizing MTL formulas.
While the exact complexity lower bounds are open, we show that the corresponding decision problem is in $\NP$.
The central SMT-based algorithm with all the theoretical results is in Section~\ref{sec:algo}.

We also implement our algorithm using a popular SMT solver in a prototype named \tool{}.
We evaluate the ability of \tool{} to synthesize $\MTL$ formulas typically employed for monitoring cyber-physical systems. 
We also empirically study the interplay between the size and future-reach of a formula. 
We present all the experimental results in Section~\ref{sec:experiments}.

\subsubsection*{Related works.} 
To our knowledge, there are only a limited number of works for synthesizing $\MTL$ formulas. 
One of them~\cite{ZheMTL} infers $\MTL$ formulas as decision trees for representing task knowledge in Reinforcement Learning.
Some other works~\cite{HoxhaDF18,YangHF12} consider the parameter search problem for $\MTL$ where, given a parametric $\MTL$ formula (i.e., an $\MTL$ formula with missing temporal bounds), they infer the ranges of parameters where the formula holds/does not hold on a given system.
Unlike our work, none of these works aims at synthesizing concise $\MTL$ specifications for monitoring tasks. 

\sloppy There are, nevertheless, numerous runtime monitoring procedures for $\MTL$~\cite{ThatiR05,BaldorN12,DokhanchiHF14,HoOW14,aerialpaper,ChattopadhyayM20,KempaZJZR20,LimaHRTY23}, clearly indicating the need for efficiently monitorable $\MTL$ specifications.
Many of them also rely on the future-reach of a specification~\cite{HoOW14,aerialpaper} or other similar measures (e.g., horizon~\cite{DokhanchiHF14}, worst-case propagation delay~\cite{KempaZJZR20}, etc.) to quantify the efficiency of their monitoring procedure.

Interestingly, several works focus on synthesizing formulas in $\STL$, an extension of $\MTL$ to reason about real-valued signals.
Bartocci et al.~\cite{BARTOCCI2022104957} provide a comprehensive survey of the existing works on inferring $\STL$.
Many of them~\cite{asarin,rpstl1,KongMTL} solve the parameter search for $\STL$, while others~\cite{dtmethod,dtbelta} learn decision trees over $\STL$ formulas, which typically do not result in concise formulas.
There are few works~\cite{MohammadinejadD20,genetic} that do prioritize the conciseness of formulas during inference.
These works cannot be directly applied to solve our problem for two main reasons.
First, these works assume inputs to be \emph{piecewise-affine continuous} signals.
While the above assumption is natural for synthesizing $\STL$ formulas inference from real-valued signals, in our setting, we must rely on the assumption that our inputs are \emph{piecewise-constant} signals, which is natural for Boolean-valued signals.
Second, these works do not employ any measure, apart from conciseness, that directly influences the efficiency of runtime monitoring.

Finally, there are works on synthesizing formulas in other temporal logics such as Linear Temporal Logic (LTL)~\cite{flie,Riener19,camachomcilraith2021,scarlet}, Property Specification Language (PSL)~\cite{0002FN20}, etc., which are not easily extensible to our setting.

%% file: prelims.tex
\label{sec:problem-setting}

\section{Preliminaries}
\label{subsec:prelims}
%In this work, we aim to infer $\MTL$ specifications from finite system observations. In~\cite{JoelMTL}, the authors describe the two commonly adapted \emph{dense-time} semantics of $\MTL$ on infinite executions: continuous and pointwise. In this study, we follow the continuous framework and extend it to finite observations inspired by~\cite{HoOW14}. Before introducing the syntax and semantics of $\MTL$, we first define the concepts of \emph{signal} and \emph{observed prefixes} representing continuous behavior of the system executions.
In this section, we introduce the basic notations used throughout the paper.
 
\paragraph{Signals and Prefixes.}
We represent continuous system executions as signals.
A \emph{signal} $\vec{x}\colon \R_{\geq 0} \to 2^\prop$ over a set of propositions $\prop$ is an infinite time series that describes relevant system events over time.
A prefix of a signal $\vec{x}$ restricted to domain $\dom = [0,T), T\in\R_{\geq 0}$ is a function $\prefix{x}\colon \dom\to 2^\prop$ where $\prefix{x}(t)=\vec{x}(t)$ for all $t\in\dom$.

To synthesize $\MTL$ formulas, we rely on finite observations that are sequences of the form $\Omega = \langle(t_i, \delta_i
)\rangle_{i\leq \finalsize}$, $\finalsize \in \N$ such that (i) $t_0 =0$, (ii) $t_{\finalsize} < T$, and (ii) for all $i \leq {\finalsize}$, $\delta_i \subseteq \prop$ is the set of propositions that hold at time point $t_i$.
%An \emph{observed signal prefix} $\prefix{x}$ is a finite sequence of tuples $\langle \left(t_i, \prefix{x}(t_i)\right) \rangle_{i \leq {\finalsize}}$ for 
To construct well-defined signal prefixes, we approximate each observation $\Omega$ as a \emph{piecewise-constant} signal prefix $\prefix{x}^{\Omega}$ using interpolation as:
%Precisely, $\prefix{x}^{\Omega}$ is constant in the interval $[t_i, t_{i+1})$ with a value of $\delta_i$ 
(i) for all $i < {\finalsize}$, for all $t \in [t_i, t_{i+1})$, $\prefix{x}(t) = \delta_i$; and (ii) for all $t \in [t_{\finalsize}, \finaltime)$, $\prefix{x}(t) = \delta_{\finalsize}$.
For brevity, we refer to signal prefixes simply as `prefixes' when clear from the context.

\paragraph{Metric Temporal Logic.} $\MTL$ is a logic formalism for specifying real-time properties of a system. 
We consider the following syntax of $\MTL$:
\[
\varphi := p \in \prop~\mid~\neg p~\mid~\varphi_1 \land \varphi_2~\mid~\varphi_1 \lor \varphi_2~\mid~\varphi_1 \lU_I \varphi_2~\mid~\lF_I \varphi~\mid~\lG_I \varphi
\]
where $p \in \prop$ is a proposition, $\neg$ is the negation operator, $\land$ and $\lor$ are the conjunction and disjunction operators respectively, and $\lU_I, \lF_I$ and $\lG_I$ are the timed-Until, timed-Finally and timed-Globally operators respectively. Here, $I$ is a closed interval of non-negative real numbers of the form $[a,b]$ where $0 \leq a \leq b$\footnote{Since we infer $\MTL$ formulas with bounded lookahead, we restrict $I$ to be bounded.}.
Note that the syntax is presented in \emph{negation normal form}, meaning that the $\neg$ operator can only appear before a proposition. 
%Also, while the full syntax of $\MTL$ includes the timed-Until ($\lU_I$) operator, we omit it in this context of inferring temporal logic specifications as the operator is often hard to interpret~\cite{KongMTL,scarlet,ZheMTL}. 

\begin{wrapfigure}{r}{0.3\textwidth}

 \centering
 \scalebox{0.9}{
    \begin{tikzpicture}
    		\node (1) at (0, 0) {$\lor$};
    		\node (2) at (-.5, -.7) {$\land$};
    		\node (3) at (.5, -.7) {$\lF_{I}$};
    		\node (4) at (0, -1.4) {$p$};
    		\node (5) at (-1, -1.4) {$\lG_{I}$};
    		\node (6) at (-1, -2.1) {$q$};
    		\draw[->] (1) -- (2); 
    		\draw[->] (1) -- (3);
    		\draw[->] (2) -- (4);
    		\draw[->] (2) -- (5);
    		\draw[->] (3) -- (4);
    		\draw[->] (5) -- (6);
    \end{tikzpicture}}
    \caption{Syntax DAG of $(p \land \lG_I q) \lor (\lF_I p)$}
\label{fig:syntax_dag}
\vspace{-5mm}
\end{wrapfigure}

As a syntactic representation of an $\MTL$ formula, we rely on \emph{syntax-DAGs}. A syntax-DAG is similar to the parse tree of a formula but with shared common subformulas. 
We define the size $|\varphi|$ of an $\MTL$ formula $\varphi$ as the number of nodes in its syntax-DAG, e.g., the size of $(p \land \lG_I q) \lor (\lF_I p)$ is six as its syntax-DAG has six nodes, as shown in Figure~\ref{fig:syntax_dag}.

As mentioned already, we follow the continuous semantics of $\MTL$. 
%For the standard continuous semantics ($\models$) of $\MTL$ over infinite signals, we refer to the work of~\cite{JoelMTL} and provide detailed descriptions in Appendix~\ref{sec:app_mtlsem}. 
First, we mention the standard continuous semantics ($\models$) of $\MTL$ over infinite signals following the work of~\cite{JoelMTL}.

Given an infinite signal $\vec{x}$, an $\MTL$ formula $\varphi$ and a time point $t \geq 0$, 

\begin{align*}
	& (\vec{x}, t) \models p   && \iff  p \in \vec{x}(t)\\
	& (\vec{x}, t) \models \neg p && \iff p \not\in \vec{x}(t)\\
	& (\vec{x}, t) \models \varphi_1 \land \varphi_2  && \iff  (\vec{x},t)\models \varphi_1\  \text{and}\  (\vec{x},t)\models \varphi_2\\
	& (\vec{x}, t) \models \varphi_1 \lor \varphi_2  && \iff  (\vec{x},t)\models \varphi_1\  \text{or}\  (\vec{x},t)\models \varphi_2\\
	& (\vec{x}, t) \models \varphi_1 \lU_{[a,b]} \varphi_2  && \iff  \exists  t' \in [t+a, t+b] \text{ s.t. }(\vec{x},t') \models \varphi_2 \text{ and }\\ & && \hspace{2cm}\forall t'' \in [t, t'], (\vec{x},t'') \models \varphi_1
\end{align*}

We read $(\vec{x},t) \models \varphi$ as `$\vec{x}$ satisfies the formula $\varphi$ at time point $t$'. The signal $\vec{x}$ satisfies the formula $\varphi$ if and only if it satisfies the formula at time point $0$, i.e., $\vec{x} \models \varphi \iff (\vec{x},0) \models \varphi$. The semantics for the $\lF_I$ and the $\lG_I$ operators can be derived using standard syntactic relations: $\lF_I \varphi := \true  \lU_I \varphi$ and $\lG_I \varphi = \neg \lF_I \varphi$.

However, our setting demands a semantics of $\MTL$ over finite prefixes such that the synthesized formulas will be `useful' while monitoring over infinite signals. Intuitively, we want an `optimistic' semantics ($\fmodels$) of an $\MTL$ formula $\varphi$ over a prefix $\prefix{x}$ such that $\prefix{x} \fmodels \varphi$ if there exists an infinite signal \emph{extending} $\prefix{x}$ that satisfies $\varphi$. In other words, $\prefix{x}$ ``carries no evidence against'' the formula $\varphi$. Formally, we want the definition of $\fmodels$ to satisfy the following lemma.
\begin{lemma}\label{lem:weak_sem}
Given a prefix $\prefix{x}$, let $ext(\prefix{x}) = \{\vec{x} \mid \prefix{x} \text{ is a prefix of } \vec{x}\}$ be the set of all infinite extensions of $\prefix{x}$. Then given an $\MTL$ formula $\varphi$, $\prefix{x} \fmodels \varphi$ if there exists $\vec{x} \in ext(\prefix{x})$ such that $\vec{x} \models \varphi$.
\end{lemma}
%The proof of the above lemma proceeds via structural induction over $\varphi$, which is describe in Appendix~\ref{sec:app_lemweaksem}.

Towards this, we follow the idea of `weak semantics' of $\MTL$ defined in~\cite{HoOW14}\footnote{Following Eisner et al.~\cite{Dana_weak}, Ho et al.~\cite{HoOW14} defined the weak semantics of $\MTL$ for the pointwise setting, which we adapt here for the continuous setting.} and interpret $\MTL$ over finite prefixes. 
Given a prefix $\prefix{x}$, we inductively define when an $\MTL$ formula $\varphi$ holds at time point $t \in \dom$, i.e., $(\prefix{x},t)\fmodels \varphi$, as follows:
% \begin{align*}
% 	& (\prefix{x}, t) \fmodels p   && \iff  p \in \prefix{x}(t) \\&(\prefix{x}, t) \fmodels \neg p  && \iff p \not\in \prefix{x}(t)\\
% 	& (\prefix{x}, t) \fmodels \varphi_1 \land \varphi_2  && \iff  (\prefix{x},t)\fmodels \varphi_1\  \text{and}\  (\prefix{x},t)\fmodels \varphi_2\\
% 	& (\prefix{x}, t) \fmodels \varphi_1 \lor \varphi_2  && \iff  (\prefix{x},t)\fmodels \varphi_1\  \text{or}\  (\prefix{x},t)\fmodels \varphi_2\\
% 	&(\prefix{x}, t)\fmodels \lF_{[a,b]}\varphi && \iff  (t +b) \geq \finaltime \text{ or } 
%     \exists  t' \in [t+a, t+b] \text{ s.t. }(\prefix{x},t') \fmodels \varphi\\
% 	&(\prefix{x}, t)\fmodels \lG_{[a,b]}\varphi && \iff  (t +a) \geq \finaltime \text{ or } \\ 
% 	&\ &&\ \hskip 0.8cm 
% 	\begin{aligned}
% 		&- \text{if $t+b <T$, }\forall  t' \in [t+a, t+b], (\prefix{x},t') \fmodels \varphi\\
% 		&- \text{otherwise, }\forall  t' \in [t+a, T), (\prefix{x},t') \fmodels \varphi
% 	\end{aligned}
% \end{align*}
\begin{align*}
	& (\prefix{x}, t) \fmodels p   \shortiff  p \in \prefix{x}(t);\\ 
        &(\prefix{x}, t) \fmodels \neg p   \shortiff p \not\in \prefix{x}(t);\\
	& (\prefix{x}, t) \fmodels \varphi_1 \land \varphi_2 \shortiff         (\prefix{x},t)\fmodels \varphi_1\  \text{and}\  (\prefix{x},t)\fmodels \varphi_2;\\
	& (\prefix{x}, t) \fmodels \varphi_1 \lor \varphi_2 \shortiff  (\prefix{x},t)\fmodels \varphi_1\  \text{or}\  (\prefix{x},t)\fmodels \varphi_2;\\
        &(\prefix{x}, t) \fmodels \varphi_1 \lU_{[a,b]} \varphi_2 \hspace{-1mm}\shortiff \\
        & \hspace{5mm} {\scriptstyle\bullet}~\exists t'\in [t+a,t+b]\cap\dom \text{ s.t. } (\prefix{x},t') \fmodels \varphi_2 \text{ and } \forall  t'' \in [t, t'], (\prefix{x},t'') \fmodels \varphi_1,~\text{or}\\
            & \hspace{5mm} {\scriptstyle\bullet}~T\leq t+b \text{ and } \forall  t'' \in [t, T),  (\prefix{x},t'') \fmodels \varphi_1\\  
	&(\prefix{x}, t)\fmodels \lF_{[a,b]}\varphi \shortiff  t +b \geq \finaltime \text{ or } 
    \exists  t' \in [t+a, t+b]\cap\dom \text{ s.t. }(\prefix{x},t') \fmodels \varphi;\\
	&(\prefix{x}, t) \fmodels \lG_{[a,b]}\varphi \shortiff t+a \geq \finaltime \text{ or } 
                \forall  t' \in [t+a, t+b] \cap \dom, (\prefix{x},t') \fmodels \varphi
\end{align*}
We say that $\prefix{x}$ satisfies $\varphi$ if $(\prefix{x},0)\fmodels \varphi$.
Also, for ensuring that our semantics complies with Lemma~\ref{lem:weak_sem}, we define $(\prefix{x}, t) \fmodels \varphi$ for all $t \geq \finaltime$ for any $\varphi$. Now, we prove that our chosen semantics satisfy the property described in Lemma~\ref{lem:weak_sem}.

\begin{proof}[Proof of Lemma~\ref{lem:weak_sem}]
	We, in fact, prove a stronger statement from which Lemma~\ref{lem:weak_sem} follows: for all $t \in [0,T), (\prefix{x}, t) \fmodels \varphi$ if there exists a signal $\vec{x} \in ext(\prefix{x})$ such that $(\vec{x},t) \models \varphi$.
	
	%Consider prefix $\prefix{x}$ and $\MTL$ formula $\varphi$ to be fixed in the proof.
	The proof now proceeds via an induction on the MTL formula $\varphi$.
	%Recall that the lemma says, given a prefix $\prefix{x}$ and an $\MTL$ formula $\varphi$, $\prefix{x} \fmodels \varphi$ if there exists an extension $\vec{x} \in ext(\prefix{x})$ such that, $\vec{x} \models \varphi$. We prove this by induction on the structure of the $\MTL$ formula $\varphi$. Fix a prefix $\prefix{x}$ and an $\MTL$ formula $\varphi$. In fact, we prove a stronger statement by induction: for all $t \in [0,T), (\prefix{x}, t) \fmodels \varphi$ if there exists a signal $\vec{x} \in ext(\prefix{x})$ such that $(\vec{x},t) \models \varphi$.
	\begin{itemize}
		\item For the base case, let $\varphi = p \in \prop$. Then, for all $t \in [0,T)$, if there exists $\vec{x} \in ext(\prefix{x})$ such that $(\vec{x},t) \models p$, then $(\prefix{x}, t) \fmodels p$ since $(\vec{x},t)\models \varphi$ and thus, $(\prefix{x},t)\models \varphi$. The same argument extends to the $\ neg$ operator.
		
		\item Let $\varphi = \varphi_1 \wedge \varphi_2$. 
		Then, for all $t\in [0,T)$, if there exists $\vec{x} \in ext(\prefix{x})$ such that $(\vec{x},t) \models \varphi_1$ and $(\vec{x},t) \models \varphi_2$.
		Then, $(\prefix{x}, t) \fmodels \varphi_1$ and $(\prefix{x}, t) \fmodels \varphi_2$ by induction hypothesis.
		The same argument extends to the $\vee$ operator.

    \item Let $\varphi = \varphi_1 \lU_{[a,b]} \varphi_2$ and fix a time point $t \in [0,T)$. We have to prove if there exists a signal $\vec{x} \in ext(\prefix{x})$ such that, $(\vec{x}, t) \models \varphi$, then $(\prefix{x},t) \fmodels  \varphi$. Now by definition of $\models$,  $\exists t' \in [t+a,t+b]$ such that, $(\vec{x}, t') \models \varphi_2$ and for all $t'' \in [t, t']$, $(\vec{x}, t') \models \varphi_1$. Now there are three cases: (i) $t+b < T$: in this case, $(\prefix{x}, t) \fmodels \varphi_1 \lU_{[a,b]} \varphi_2$ by definition of $\fmodels$, (ii) $T \leq t' \leq t+b$: in this case, $\forall t'' \in [t,T), (\prefix{x}, t'') \fmodels \varphi_1$ and hence, $(\prefix{x}, t'') \fmodels \varphi$, and (iii) $t' <T \leq t+b$: this case is similar to the first case. 
\end{itemize}
	The cases for $\varphi = \lF_{[a,b]} \psi$ and $\varphi = \lG_{[a,b]} \psi$ can be proved similarly using case analysis. \qed
\end{proof}

\section{The Problem Formulation}\label{subsec:problem-form}
Next, we formally introduce the various aspects of the central problem of the paper.

\paragraph{Sample.}
The input data consists of a set of labeled (piecewise-constant) prefixes. 
Formally, we rely on a sample $\mathcal{S}=(P,N)$ consisting of a set $P$ of positive prefixes and a set $N$ of negative prefixes such that $P\cap N = \emptyset$.
We say an $\MTL$ formula $\varphi$ is \emph{globally-separating} ($\gsep$, for short) for $\mathcal{S}$ if it satisfies all the positive prefixes at each time point and does not satisfy negative prefixes at some time point\footnote{Most stream-based monitors check if the specification holds at every time point~\cite{aerialtool}.}.
Formally, given a sample $\sample$, we define an $\MTL$ formula $\varphi$ to be $\gsep$ for $\sample$ if (i) for all $\prefix{x} \in P$ and for all $t \in [0,T)$, $(\prefix{x},t)  \fmodels \varphi$; and 
(ii) for all $\prefix{y} \in N$, there exists $t \in [0,T)$ such that $(\prefix{y}, t) \not \fmodels \varphi$.

\paragraph{Future-Reach.} \label{sec:fr}
%\emph{Monitoring} of an $\MTL$ specification against an infinite execution of a system, represented as a signal, refers to the process of continuously checking whether the signal conforms to the specification at each time point. In the `online' setting, this check happens in real-time, i.e., the (streaming) signal is given incrementally over time. Hence, the complexity of this procedure not only depends on the conciseness of the specification but also on the amount of `lookahead' needed to check the satisfaction of the specification at a given time point. 
%Previously, we have explained how the `lookahead' of an $\MTL$ formula has an impact on the efficiency of an online monitor monitoring the formula.
To formalize the lookahead of an $\MTL$ formula $\varphi$, we rely on its future-reach $\fr(\varphi)$, following~\cite{hunter,HoOW14}, which indicates how much of the future is required to determine the satisfaction of $\varphi$. It is defined inductively as follows: 
\begin{align*}
&\fr(p) = \fr(\neg p) = 0\\
&\fr(\varphi_1 \land \varphi_2) = \fr(\varphi_1 \lor \varphi_2) =  \max(\fr(\varphi_1), \fr(\varphi_2))\\
&\fr(\varphi_1 \lU_{[a,b]} \varphi_2) = b + \max(\fr(\varphi_1), \fr(\varphi_2))\\
&\fr( \lF_{[a,b]} \varphi) = \fr(\lG_{[a,b]} \varphi) =  b + \fr(\varphi)
\end{align*}

To highlight that $\fr(\varphi)$ quantifies the lookahead of $\varphi$, we observe the following lemma: 
% \begin{lemma}\label{lem:fr-property}
%     Let $\varphi$ be an $\MTL$ formula such that $\fr(\varphi) \le k$.
%     Also, let $\prefix{x}$ and $\prefix{y}$ be two prefixes such that $k\in\dom$ and $\vec{x}_{[0,k]}=\vec{y}_{[0,k]}$.
%     Then, $\prefix{x} \fmodels \varphi$ if and only if $\prefix{y} \fmodels \varphi$.
% \end{lemma}
% \todo{stronger}
\begin{lemma}\label{lem:fr-property}
    Let $\varphi$ be an $\MTL$ formula such that $\fr(\varphi) \le K$ for some $K \in \R^{\geq 0}$. Let $\vec{x}$ and $\vec{y}$ be two signals such that $\vec{x}_{[0,K]}=\vec{y}_{[0,K]}$.
    Then, for all $T \in \R^{\geq 0}$, $\prefix{x} \fmodels \varphi$ if and only if $\prefix{y} \fmodels \varphi$.
\end{lemma}

Intuitively, the above lemma states that a formula with future-reach $\le K$ cannot distinguish between two signals that are identical up to time $K$. We prove the lemma below.
%The lemma can be proved using structural induction over $\varphi$. For space constraints, we include the whole proof in Appendix~\ref{sec:app_frlem}.
\begin{proof}[Proof of Lemma~\ref{lem:fr-property}]
	We will prove this by induction on the structure of $\varphi$. In particular, we will prove the following:
	
	For any $K$, let $\varphi$ be a formula with $\fr(\varphi) \le K$ and $\vec{x}$ and $\vec{y}$ be two signals such that $\vec{x}_{[0,K]}=\vec{y}_{[0,K]}$. Then, for all $T \in \R^{\geq 0}$, $\prefix{x} \fmodels \varphi$ if and only if $\prefix{y} \fmodels \varphi$.
	
	- For the base case, let $\varphi = p$. Then, $p \in \prefix{x}(0)$ and as $\vec{x}_{[0,K]}=\vec{y}_{[0,K]}$, $p \in \prefix{y}(0)$. Hence, $\prefix{y} \fmodels p$. This can be similarly seen for the case where $\varphi = \neg p$.
	
	- The proof for the cases where $\varphi = \varphi_1 \lor \varphi_2$ or $\varphi = \varphi_1 \land \varphi_2$ can be derived easily.
	
	- Let $\varphi = \lF_{[a,b]} \varphi_1$. Let us fix a $T$ such that $\prefix{x} \fmodels \lF_{[a,b]}\varphi$. If $b \geq t$, then $\prefix{y} \fmodels \lF_{[a,b]}\varphi$ trivially. If not, then there exists a time point $t \in [a,b]$ such that $(\prefix{x},t) \fmodels \varphi_1$. Now, let $\vec{x'} = \vec{x}^{[t:]}$ and $\vec{y'} = \vec{y}^{[t:]}$ be the signals obtained by shifting the original signals by $-t$. Formally, $\forall t' \in \R^{\geq 0}$, $\vec{x'}(t') = \vec{x}(t'+t)$ and $\vec{y'}(t') = \vec{y}(t'+t)$. Note that, $\vec{x'}_{[0, K-t]} = \vec{y'}_{[0, K-t]}$. Also, $\fr(\varphi_1) = \fr (\varphi) - b \leq K-b \leq K-t$ and $\vec{x'}_{[0, K-t]} \fmodels \varphi$. Then, following induction hypothesis, $\vec{y'}_{[0, K-t]} \fmodels \varphi_1$ which implies that $(\prefix{y}, t) \fmodels \varphi_1$. Hence, $\prefix{y} \fmodels \lF_{[a,b]} \varphi$. The case where $\varphi = \lG_{[a,b]} \varphi_1$ can be proved similarly.
	
	- Let $\varphi = \varphi_1 \lU_{[a,b]} \varphi_2$. Again, similar to above, fix a $T$ such that $\prefix{x} \fmodels \varphi_1 \lU_{[a,b]}\varphi_2$. Let us first assume that $b \le T$. Then, $\exists t \in [a,b]$ such that $(\prefix{x},t) \fmodels \varphi_2$ and $\forall t' \in [0,t]$, $(\prefix{x},t') \fmodels \varphi_1$. Now as $\fr(\varphi_1)$ and $\fr(\varphi_2)$ are both $\le K - b \le K- t$. Hence again using similar methods as above, one can prove that $(\prefix{y},t) \fmodels \varphi_2$ and $\forall t' \in [0,t]$, $(\prefix{y},t') \fmodels \varphi_1$. Hence, $\prefix{y} \fmodels \varphi_1 \lU_{[a,b]}\varphi_2$.
\end{proof}

\paragraph{The Problem.}
We now formally introduce the problem of synthesizing an $\MTL$ formula.
In the problem, we ensure that the $\MTL$ formula is efficient for monitoring by allowing the system designer to specify a future-reach bound. %We formalize this $\MTL$ synthesis problem as follows:
%Thus, to formally identify the specification we synthesize, we introduce the term $(\sample,\bound)-\sep$ for an $\MTL$ formula. 
%We say an $\MTL$ formula $\varphi$ is a $(\sample,\bound){-}\sep$ if 
%\begin{enumerate}
%\item  $\varphi$ is invariant for the given sample $\sample$;
%\item $\fr(\varphi)$ is less than $\bound$;
%\end{enumerate}
%When clear from the context, we omit the parameters $(\sample,\bound)$ to describe a $\sep$. We are now ready to present the learning problem.
%\noindent
\begin{problem}[\syn] \label{problem:main}
	Given a sample $\sample = (P,N)$ and a future-reach bound $\bound$, find an $\MTL$ formula $\varphi$ such that (i) $\varphi$ is $\gsep$ for $\sample$; (ii) $\fr(\varphi) \leq \bound$; (iii) for every $\MTL$ formula $\varphi'$ such that $\varphi'$ is $\gsep$ for $\sample$ and $\fr(\varphi')\leq\bound$, $|\varphi|\leq|\varphi'|$.
\end{problem}
Intuitively, the above optimization problem asks to synthesize a minimal size $\MTL$ formula that is $\gsep$ for the input sample and has a future-reach within the input bound. 
Before we dive into the procedure for finding such an MTL formula, we first study if such an MTL formula even exists.

\section{Existence of a solution}
\label{sec:existence}

As alluded to in the introduction, for any given sample $\sample$ and future-reach bound $K$, the existence of a suitable $\lG$-separating formula is not always guaranteed.
For an illustration, consider the sample $\sample$ with one positive prefix $\prefix{x} = \langle (0,\{q\}),(2,\{\}) \rangle$ and one negative prefix $\prefix{y}=\langle (0,\{q\}) \rangle$, and domain $\dom=[0,4)$.
For $\sample$, there is no formula $\varphi$ with $\fr(\varphi)\leq 1$ that is $\gsep$.
To see this, assume there exists a prospective formula $\varphi$.
Consequently, $\varphi$ being $\gsep$, $(\prefix{x},0)\models\varphi$.
Observe that, for all time-points $t\in \dom$, $\prefix{y}$ when restricted to time interval $[t,t+1]\cap \dom$ appears identical to $\prefix{x}$ when restricted to time interval $[0,1]$ to $\varphi$ since its future-reach is 1 (using Lemma~\ref{lem:fr-property}).
Thus, for all time-points $t\in\dom$, $(\prefix{y},t)\models\varphi$ violating that $\varphi$ is $\gsep$.

What we show now is that one can check whether a prospective formula exists by relying on a simple characterization of the inputs $\sample$ and $K$.
Towards this, we introduce introduce some terminology.

We introduce the infix of a prefix $\prefix{x}$ that is a restriction of $\prefix{x}$ to a specific time interval.
Formally, given two time-points $t_1\leq t_2<\finaltime$ and a prefix $\prefix{x}$, infix $\prefix{x}^{[t_1,t_2]}$ is the function $\prefix{x}^{[t_1,t_2]}\colon [0,t_2-t_1]\to 2^\prop$ such that $\prefix{x}^{[t_1,t_2]}(t)=\prefix{x}(t+t_1)$ for all $t\in[0,t_2-t_1]$.

%Next, we say that a prefix $\prefix{x}$ is $\gksepable$ from a prefix $\prefix{y}$ if there exists a two time points $t_1\leq t_2<\finaltime$ with $t_2-t_1\leq K$ such that $\prefix{y}^{[t_1,t_2]}\neq \prefix{x}^{[t^{'}_1,t^{'}_2]}$ for any infix $\prefix{x}^{[t^{'}_1,t^{'}_2]}$ of $\prefix{x}$.
%In order words, $\prefix{x}$ is $\gksepable$ from $\prefix{y}$ if there exists an infix $\prefix{y}^{[t_1,t_2]}$ with $t_2-t_1\leq K$ that is not an infix of $\prefix{x}$.

Next, we define a characterization of a sample $\sample$ based on the future-reach $K$, which we term as $\gksepable$.
Intuitively, we say $\sample$ to be $\gksepable$ if there is a $K$-length infix $\prefix{y}^{[t_1,t_2]}$ for every negative prefix $\prefix{y}$ in $\sample$ that is not an infix of any positive prefix in $\sample$.
Formally, $\sample=(P,N)$ is $\gksepable$ if 
for every negative prefix $\prefix{y}\in N$, there exists an infix $\prefix{y}^{[t_1,t_2]}$ with $t_2-t_1\leq K$ such that $\prefix{y}^{[t_1,t_2]}\neq \prefix{x}^{[t^{'}_1,t^{'}_2]}$ for any infix $\prefix{x}^{[t^{'}_1,t^{'}_2]}$ of any positive prefix $\prefix{x}\in P$.
%$\prefix{x}$ is globally-separable with bound $\bound$ from $\prefix{y}$ for every positive prefix $\prefix{x}\in P$ and every negative prefix $\prefix{y}\in N$.

We now state the result that enables checking the existence of a solution to Problem~\ref{problem:main}.
\begin{lemma}
For a given sample $\sample$ and future-reach bound $\bound$, there exists an MTL formula $\varphi$ with $\fr(\varphi) \leq \bound$ that is $\gsep$ for $\sample$ if and only if $\sample$ is $\gksepable$. 
\end{lemma}
\begin{proof}
($\Rightarrow$) For the forward direction, consider $\varphi$ be an MTL formula with $\fr(\varphi)\leq \bound$ that is $\gsep$ for $\sample$.
% For a proof by contradiction, assume that $\sample$ is not $\gksepable$.
% This means that there exists some negative prefix, say  $\prefix{\bar{y}}\in N$, for which the following holds: for all of its infixes $\prefix{\bar{y}}^{[t_1,t_2]}$ where $t_2-t_1\leq K$, there is some infix $\prefix{\bar{x}}^{[t^{'}_1,t^{'}_2]}$ of some positive prefix $\prefix{\bar{x}}\in P$ such that $\prefix{\bar{x}}^{[t^{'}_1,t^{'}_2]}=\prefix{\bar{y}}^{[t_1,t_2]}$.
Since $\varphi$ is $\gsep$, for any arbitrary negative prefix, say $\prefix{\bar{y}}$, there must be a time-point, say $\bar{t}<T$, such that $(\prefix{\bar{y}},\bar{t})\not\models \varphi$.
If $\bar{t}+K<T$, we show by contradiction that the infix $\prefix{\bar{y}}^{[\bar{t},\bar{t}+K]}$ is not an infix in any positive prefix.
In particular, if $\prefix{\bar{y}}^{[\bar{t},\bar{t}+K]} = \prefix{x}^{[t,t+K]}$, then $(\prefix{x},t)\not\models\varphi$ as $\varphi$ cannot distinguish between signals that are identical up to time $K$ (using Lemma~\ref{lem:fr-property}).
If $\bar{t}+K\geq T$, the semantics of MTL being weak, there is an $L<K$ with $\bar{t}+L<T$ such that
for any $\vec{y}\in ext(\prefix{y}^{[0,\bar{t}+L]})$, $(\vec{y},t)\not\models \varphi$ (using Lemma~\ref{lem:weak_sem}).
Once again, we show by contradiction that the infix $\prefix{\bar{y}}^{[\bar{t},\bar{t}+L]}$ is not an infix in any positive prefix.
In particular, if $\prefix{\bar{y}}^{[\bar{t},\bar{t}+L]} = \prefix{x}^{[t,t+L]}$, then for all $\vec{x}\in ext(\prefix{x}^{[0,t+L]})$ 
$(\vec{x},t)\not\models\varphi$. 
Also, for any $\vec{x}\in ext(\prefix{x})$ 
$(\vec{x},t)\not\models\varphi$, meaning $(\prefix{x},t)\not\models \varphi$ (again, using Lemma~\ref{lem:weak_sem}). 

($\Leftarrow$) For the other direction, consider $\sample$ to be $\gksepable$.
Using the definition of $\gksepable$, for any arbitrary negative prefix, say $\prefix{\bar{y}}$, we have an infix $\prefix{\bar{y}}^{[t_1,t_2]}$ with $t_2-t_1\leq K$ that is not an infix in any positive prefix.
We construct a formula $\varphi_{\prefix{\bar{y}}}$ that explicitly specifies the propositions appearing in each interval of the infix $\prefix{\bar{y}}^{[t_1,t_2]}$ using $\lG$ and  $\wedge$ operators.
Observe that $\fr(\varphi_{\prefix{\bar{y}}})\leq K$ since $t_2-t_1\leq K$ in $\prefix{\bar{y}}^{[t_1,t_2]}$.
Now, the formula $\neg \varphi_{\prefix{x}}$ holds at all time-points in all positive prefixes, while it does not hold at time-point $t_1$ in $\prefix{\bar{y}}$.
We finally construct the prospective formula as $\varphi = \bigwedge_{\prefix{y}\in N}\neg\varphi_{\prefix{y}}$ which is $\gsep$ for $\sample$ and also, $\fr(\varphi)\leq K$.
\qed

\end{proof}

%Let $T>1$ and the sample $\sample$ consists of one positive prefix: $(0, \{p\}) , (1, \{q\})$ and one negative prefix: $(0, \{q\})$. One can check that there does not exist any $\MTL$ formula, which is $\gsep$ for $\sample$. 
% Hence, to ensure that our algorithm terminates and we obtain a concise specification, we impose a reasonable size bound $B$ up to which we search for a solution. 
%Also, note that, in this problem, our goal is to find only one $\MTL$ formula, although there might be multiple solutions.

%note that the solution may not be unique, and our goal is to find one such formula if multiple exist.

We now describe an $\NP$ algorithm to check whether a sample $\sample$ is $\gksepable$.
The crux of the algorithm is to guess, for each negative prefix $\prefix{y}$, an infix $\prefix{y}^{[t_1,t_2]}$ with $t_2-t_1\leq K$ and then check whether it is an infix of any positive prefix.
The procedure of checking involves comparing the various intervals of $\prefix{y}^{[t_1,t_2]}$ against the intervals of infixes of positive prefixes.

To describe the checking procedure in detail, let $\prefix{\bar{y}}^{[t_1,t_2]}$ be an infix of the negative prefix $\prefix{\bar{y}}$. 
We like to check whether $\prefix{\bar{y}}^{[t_1,t_2]}$ is an infix of the positive prefix $\prefix{\bar{x}}$.
To do so, we check $\prefix{\bar{y}}^{[t_1,t_2]}=\prefix{\bar{x}}^{[t,t+(t_2-t_1)]}$ with only those infixes in which the time-points where $\prefix{x}$ and $\prefix{y}$ have been observed coincide. 
Precisely, we check $\prefix{\bar{y}}^{[t_1,t_2]} = \prefix{\bar{x}}^{[t,t+(t_2-t_1)]}$ for all those infixes of $\prefix{\bar{x}}$ where $t''-t = t'-t_1$, $t''$ and $t'$ being timepoints where $\prefix{x}$ and $\prefix{y}$ have been observed, respectively.
This process is based on the fact that the changes in an infix occur only at the observation time points.
Also, this process takes time polynomial in the number of observation time-points of $\prefix{\bar{x}}$ and $\prefix{\bar{y}}$.
We can perform the procedure for each positive and negative prefix.
Overall, we have the following result.

\begin{lemma}
Given a sample $\sample$ and future-reach bound $K$, checking whether $\sample$ is $\gksepable$ can be done in $\NP$.
\end{lemma}

%% file: algorithms.tex
\section{An SMT-based Algorithm}\label{sec:algo}
Our algorithm relies on an SMT-based approach inspired by the numerous constraint satisfaction-based approaches for synthesizing temporal logic formulas~\cite{flie,CamachoM19,0002FN20,ArifLERCT20}.
Roughly speaking, our algorithm constructs a series of formulas in Linear Real Arithmetic (LRA) and uses an optimized SMT solver to search for the desired solution.
To expand on the specifics of our algorithm, we first familiarize the readers with LRA.

\paragraph{Linear Real Arithmetic (LRA).}
In LRA~\cite{BarrettSST21}, given a set of real variables $\mathcal{Y}$, a \emph{term} is defined recursively as either constant $c \in \R$, a real variable $y \in \mathcal{Y}$, 
a product $c\cdot y$ of a constant $c
\in\R$ and a real variable $y \in \mathcal{Y}$, or a sum $t_1 + t_2$ of two terms $t_1$ and $t_2$.
An \emph{atomic formula} is of the form $t_1\diamond t_2$ where $\diamond\in\{<,\leq,=,\geq,>\}$.
An \emph{LRA formula}, defined recursively, is either an atomic formula, the negation $\lnot \Phi$ of an LRA formula $\Phi$, or the disjunction $\Phi_1 \lor \Phi_2$ of two formulas $\Phi_1, \Phi_2$.
We additionally include standard Boolean constants $\true$, and $\false$ and Boolean operators $\land$, $\rightarrow$ and $\leftrightarrow$. 

To assign meaning to an LRA formula, we rely on a so-called \emph{interpretation} function $\inter \colon \mathcal{Y} \to \R$ that maps real variables to constants in $\R$.
An interpretation $\inter$ can easily be lifted to a term $t$ in the usual way, and is denoted by $\inter(t)$.
We now define when $\inter$ \emph{satisfies} a formula $\varphi$, denoted by $\inter \models \varphi$, recursively as follows: $\inter \models t_1 \diamond t_2$ for $\diamond \in \{ <, \leq, =, \geq, > \}$ if and only if $\inter(t_1) \diamond \inter(t_2)$ is $\true$, $\inter \models \lnot \Phi$ if $\inter \not\models \Phi$, and $\inter \models \Phi_1 \lor \Phi_2$ if and only if $\inter \models \Phi_1$ or $\inter \models \Phi_2$.
We say that an LRA formula $\Phi$ is \emph{satisfiable} if there exists an interpretation $\inter$ with $\inter \models \Phi$.

Despite being \NP-complete, with the rise of the SAT/SMT revolution~\cite{sat-2022}, checking the satisfiability of LRA formulas can be handled effectively by several highly-optimized SMT solvers~\cite{MouraB11,CimattiGSS13,BarbosaBBKLMMMN22}.

\subsubsection{Algorithm Overview.}
Our algorithm constructs a series of LRA formulas $\langle\Phi^n_{\sample,\bound}\rangle_{n=1,2,\ldots}$ to facilitate the search for a suitable $\MTL$ formula. 
The formula $\Phi^n_{\sample,\bound}$ has the following properties:
\begin{enumerate}
	\item $\Phi^n_{\sample,\bound}$ is satisfiable if and only if there exists an $\MTL$ formula $\varphi$ of size $n$ such that $\varphi$ is $\gsep$ for $\sample$ and $\fr(\varphi)\leq \bound$. 
	\item from any satisfying interpretation $\inter$ of $\Phi^n_{\sample,\bound}$, one can construct an appropriate $\MTL$ formula $\varphi^\inter$.
\end{enumerate} 

In our algorithm, sketched in Algorithm~\ref{alg:main}, we first check whether 
$\sample$ is $\gksepable$ (as described in Section~\ref{sec:existence}) which informs us whether a prospective formula exists.
We now check the satisfiability of $\Phi^n_{\sample,\bound}$ for increasing values of size $n$ starting from 1.
If $\Phi^n_{\sample,\bound}$ is satisfiable for some $n$, then our algorithm constructs a prospective $\MTL$ formula $\varphi^\inter$ from a satisfying interpretation $\inter$ returned by the SMT solver.
This algorithm terminates because of checking whether a solution exists apriori and it returns a minimal formula because of the iterative search through MTL formulas of increasing sizes.

\begin{algorithm}[t]
	\caption{Overview of our algorithm}
	\label{alg:main}
	\begin{algorithmic}[1]
		\Statex	\textbf{Input:} Sample $\sample$, $\fr$-bound $K$
            
            \If{$\sample$ is not $\gksepable$}
            \Return No prospective formula 
            \EndIf
	\State $n\gets 0$	
        \While{True}
            \State $n\gets n+1$
		\State Construct $\Phi^{n}_{\sample,K} \coloneqq \Phi^{str}_{n,\sample,K} \wedge \Phi^{\fr}_{n,\sample,K} \wedge \Phi^{sem}_{n,\sample,K}$
		\If{$\Phi^n_{\sample,K}$ is SAT}
		\State Construct $\varphi^\inter$ from a satisfying interpretation $\inter$
            \Return $\varphi^{\inter}$
		\EndIf
		\EndWhile
	\end{algorithmic}
\end{algorithm}

The crux of our algorithm lies in the construction of the formula $\Phi^n_{\sample,\bound}$. 
Internally, $\fullcons \coloneqq \strcons_{n,\sample,\bound} \wedge \frcons_{n,\sample,\bound} \wedge \semcons_{n,\sample,\bound}$ is a conjunction of three subformulas, each with a distinct role.
The subformula $\strcons_{n,\sample,\bound}$ encodes the structure of the prospective $\MTL$ formula.
The subformula $\frcons_{n,\sample,\bound}$ ensures that the future-reach of the prospective formula is less than or equal to $\bound$. 
Finally, the subformula $\semcons_{n,\sample,\bound}$ ensures that the prospective formula is $\gsep$ for $\sample$.
In what follows, we expand on the construction of each of the introduced subformulas.
We drop the subscripts $n$, $\sample$, and $\bound$ from the subformulas when clear from the context.

\paragraph{Structural Constraints.}
Following Neider and Gavran~\cite{flie}, we symbolically encode the syntax-DAG of the prospective $\MTL$ formula using the formula $\strcons$.
For this, we first fix a naming convention for the nodes of the syntax-DAG of an $\MTL$ formula. 
For a formula of size $n$, we assign to each of its nodes an identifier from $\{1,\dots, n\}$ such that the identifier of each node is larger than that of its children if it has any.
Note that such a naming convention may not be unique.
Based on these identifiers, we denote the subformula of $\varphi$ rooted at Node~$i$ as $\varphi[i]$.
In that case, $\varphi[n]$ is precisely the formula $\varphi$.

Next, to encode a syntax-DAG symbolically, we introduce the following variables\footnote{We include Boolean variables in our LRA formulas since Boolean variables can always be simulated using real variables that are constrained to be either 0 or 1.}: (i) Boolean variables $x_{i,\lambda}$ for $i\in\{1,\dots, n\}$ and $\lambda\in\prop\cup\{\neg,\lor,\land,\lU_I,\lF_I,\lG_I\}$; (ii) Boolean variables $l_{i,j}$ and $r_{i,j}$ for $i\in\{1,\dots,n\}$ and $j\in\{1,\dots,i\}$; (iii) real variables $a_{i}$ and $b_{i}$ for $i\in\{1,\dots,n\}$.
The variable $x_{i,\lambda}$ tracks the operator labeled in Node~$i$, meaning, $x_{i,\lambda}$ is set to true if and only if Node~$i$ is labeled with $\lambda$.
The variable $l_{i,j}$ (resp., $r_{i,j}$) tracks the left (resp., right) child of Node~$i$, meaning, $l_{i,j}$ (resp., $r_{i,j}$) is set to true if and only if the left (resp., right) child of Node~$i$ is Node~$j$.
Finally, the variable $a_i$ (resp., $b_i$) tracks the lower (resp., upper) bound of the interval $I$ of a temporal operator (i.e., operators $\lU_I$, $\lF_I$ and $\lG_I$), meaning that, if $a_i$ (resp. $b_i$) is set to $a\in\R$ (resp., $b\in\R$), then the lower (resp., upper) bound of the interval of the operator in Node~$i$ is $a$ (resp., $b$).
While we introduce variables $a_i$ and $b_i$ for each node, they become relevant only for the nodes that are labeled with a temporal operator.

We now impose structural constraints on the introduced variables to ensure they encode valid $\MTL$ formulas.
%Exemplarily, we have the following constraint:
%\[
%\Big[ \bigwedge_{1\leq i \leq n} \bigvee_{\lambda \in \Lambda} x_{i,\lambda} \Big] \land \Big[\bigwedge_{1 \leq i \leq n} \bigwedge_{\lambda \neq \lambda' \in \Lambda} \lnot x_{i,\lambda} \lor \lnot x_{i,\lambda'}  \Big],
%\]
%where $\Lambda=\prop\cup\{\neg,\lor,\land,\lU_I,\lF_I,\lG_I\}$.
%The above constraint ensures that each node is labeled by exactly one operator or one proposition.
%We also impose other structural constraints, such as each node can have at most two children, Node~$1$ must be a proposition, etc.
These constraints are similar to the ones proposed by Neider and Gavran~\cite{flie}.
%We here additionally ensure that the $\neg$ operator appears only in front of propositions.
%Also, we ensure that the intervals of the temporal operators are proper using the constraint $\bigwedge_{1\leq i\leq n} 0\leq a_i \leq b_i < \bound$.
%We refer interested readers to Appendix~\ref{sec:app_structur} for all the constraints.
For each Node~$i$ containing operator $\lambda$, we define the following two functions:
\begin{equation*}
	\begin{aligned}
		&\mathit{exactly-one-left}(i) = \Big[\bigwedge\limits_{1\leq i\leq  n} \bigvee\limits_{1\leq j\leq i} l_{i,j}] \wedge[\bigwedge\limits_{2\leq i\leq  n}\bigwedge\limits_{1\leq j\leq j'\leq n}\neg l_{i,j}\vee \neg l_{i,j'}\Big] \text{, and}\\
		&\mathit{exactly-one-right}(i) = \Big[\bigwedge\limits_{1\leq i\leq  n} \bigvee\limits_{1\leq j\leq i} r_{i,j}] \wedge[\bigwedge\limits_{2\leq i\leq  n}\bigwedge\limits_{1\leq j\leq j'\leq n}\neg r_{i,j}\vee \neg r_{i,j'}\Big]
	\end{aligned}
\end{equation*}
that defines that the node contains exactly one left child and exactly one right child, respectively.

Now let $\Lambda = \mathcal{P} \cup U_\Lambda \cup B_\Lambda$, where $U_\Lambda$ denotes the set of unary operators and $B_\Lambda$ denotes the set of binary operators. Then the encoding of the structural constraints contains the following:

\begin{align}
	&\Big[ \bigwedge_{1\leq i \leq n} \bigvee_{\lambda \in \Lambda} x_{i,\lambda} \Big] \land \Big[\bigwedge_{1 \leq i \leq n} \bigwedge_{\lambda \neq \lambda' \in \Lambda} \lnot x_{i,\lambda} \lor \lnot x_{i,\lambda'}  \Big] \land \label{cons:one_label}\\
	&\bigwedge_{1\leq i \leq n} \left(\bigvee_{p \in \mathcal{P}} x_{i,p} \rightarrow \Big[\bigwedge_{1 \leq j \leq n} \neg l_{i,j} \land \bigwedge_{1 \leq j \leq n} \neg r_{i,j'} \Big]\right) \land \label{cons:prop}\\
	&\bigwedge_{1\leq i \leq n} \left(\bigvee_{\lambda \in U_\Lambda} x_{i,\lambda} \rightarrow \Big[\mathit{exactly-one-left}(i) \land \bigwedge_{1 \leq j \leq n} \neg r_{i,j'} \Big]\right) \land \label{cons:unary}\\
	&\bigwedge_{1\leq i \leq n} \left(\bigvee_{\lambda \in B_\Lambda} x_{i,\lambda} \rightarrow \Big[\mathit{exactly-one-left}(i) \land \mathit{exactly-one-right}(i) \Big]\right) \label{cons:binary}\\
	&\bigwedge_{\substack{{1\leq i\leq n}\\{1\leq j < i}}} x_{i,\neg} \wedge l_{i,j} \rightarrow \big[\bigvee_{p\in\prop} x_{j,p}\big] \label{cons:nnf}
\end{align}

Constraint~\ref{cons:one_label} encodes that each node only contains one operator or proposition. Constraint~\ref{cons:prop} imposes that the nodes containing a proposition do not have any child. Constraint~\ref{cons:unary} says that the nodes containing a unary operator contain exactly one child, while constraint~\ref{cons:binary} enforces that the nodes containing a binary operator contain exactly one left and exactly one right child.
Finally, Constraint~\ref{cons:nnf} imposes that the $\ neg$ operator can occur only in front of propositions.

The subformula $\strcons$ is a conjunction of all the structural constraints we described.
Using a satisfying interpretation $\inter$ of $\strcons$, one can construct the syntax DAG of a unique $\MTL$ formula $\varphi^\inter$.

\paragraph{Future-reach Constraints.}
To symbolically compute the future-reach of the prospective formula $\varphi$, we encode the inductive definition of the future-reach, as described in Section~\ref{sec:fr} in an LRA formula. 
To this end, we introduce real variables $f_i$ for $i \in \{1,\dots,n\}$ to encode the future-reach of the subformula $\varphi[i]$.
Precisely, $f_i$ is set to $f\in\R$ if and only if $\fr(\varphi[i])=f$. 

To ensure the desired meaning of the $f_i$ variables, we impose the following constraints:

\begin{align*}
	&\bigwedge_{1\leq i\leq n} x_{i,p}  \rightarrow \big[f_i=0\big] \land
	\bigwedge_{\substack{{1\leq i\leq n}\\{1\leq j < i}}}\left(x_{i,\neg} \wedge l_{i,j}\right) \rightarrow \big[f_i=f_j\big] \land\\
 &\bigwedge_{\substack{{1\leq i\leq n}\\{1\leq j,j' < i}}}\left( \left( x_{i,\lor} \lor x_{i,\land}\right)  \wedge l_{i,j} \wedge r_{i,j'}\right)  \rightarrow \big[f_i=\max(f_j, f_j')\big] \land\\
	&\bigwedge_{\substack{{1\leq i\leq n}\\{1\leq j < i}}}\left( x_{i,\lF_I} \wedge l_{i,j} \right)\rightarrow \big[f_i=f_j+b_i\big] \land
	\bigwedge_{\substack{{1\leq i\leq n}\\{1\leq j < i}}}\left( x_{i,\lG_I} \wedge l_{i,j} \right)\rightarrow \big[f_i=f_j+b_i\big]
\end{align*}
Each line above imposes constraints based on the definition of future-reach for that operator, described in Section~\ref{sec:fr}. 
%
% such as
%\begin{align}
%\bigwedge_{{1\leq i\leq n},{1\leq j < i}} [x_{i,\lF_I} \wedge l_{i,j}] \rightarrow [f_i=f_j+b_i].
%\end{align}
%This constraint expresses that if Node~$i$ contains the $\lF_I$ operator where $I$ is encoded using $a_i$ and $b_i$, then the future-reach of $\varphi[i]$, i.e., $\fr(\varphi[i])$), must be the future-reach of $\varphi[j]$ plus $b$, i.e., $\fr(\varphi[j])+b$.
%For other operators, we impose similar 
%We refer the readers to Appendix~\ref{sec: app_fr} for the remaining future-reach constraints.

Finally, to enforce that the future-reach of the prospective $\MTL$ formula is within $\bound$, along with the constraints mentioned above, we have $f_n \leq \bound$ in $\frcons$.

\paragraph{Semantic Constraints.} To symbolically check whether the prospective formula is $\gsep$, we must encode the procedure of checking the satisfaction of an $\MTL$ formula into an LRA formula.
To this end, we rely on the monitoring procedure devised by Maler and Nickovic~\cite{maler} for efficiently checking when a signal satisfies an $\MTL$ formula.
Since our setting is slightly different, we take a brief detour via the description of our adaptation of the monitoring algorithm.

Given an $\MTL$ formula $\varphi$ and a signal prefix $\prefix{x}$, our monitoring algorithm computes the (lexicographically) ordered set $\ivalset{}{\varphi} = \{I_1,\cdots,I_\eta\}$ of \emph{maximal disjoint time intervals} $I_1, \cdots, I_\eta$ where $\varphi$ holds on $\prefix{x}$.
Mathematically speaking, the following property holds for the set $\ivalset{}{\varphi}$ we construct:
\begin{lemma}\label{lem:corr_maler}
	Given an $\MTL$ formula $\varphi$ and a prefix $\prefix{x}$, for all $t \in \dom$, $(\prefix{x},t) \fmodels \varphi$ if and only if $t \in I$ for some $I\in \ivalset{}{\varphi}$.
\end{lemma}

In our monitoring algorithm, we compute the set $\ivalset{}{\varphi}$ inductively on the structure of the formula $\varphi$.
To describe the induction, we use the notation $\ivalset{\cup}{\varphi} = \bigcup_{I\in \ivalset{}{\varphi}} I$ to denote the union of the intervals in $\ivalset{}{\varphi}$.
For the base case, we compute $\ivalset{}{p}$ for every $p\in\prop$ by accumulating the time points $t\in[0,T)$ where $(\prefix{x},t)\fmodels p$ into maximal disjoint time intervals.
In the inductive step, we exploit the relations presented in Table~\ref{tab:monitoring-algo} for the different $\MTL$ operators.
In the table, $[t_1,t_2)\ominus[a,b] = [t_1-b,t_2-a)\cap \dom$ and $\mathcal{I}^{c} = \dom - \I$.
\begin{table}[h]
\vspace{-3mm}
\caption{The relations for inductive computation of $\ivalset{\cup}{\varphi}$.}\label{tab:monitoring-algo}
\vspace{2mm}
\centering
\renewcommand{\arraystretch}{1.4}
\begin{tabular}{|c|}
\hline
$\ivalset{\cup}{\neg p} = \left(\ivalset{\cup}{p}\right)^{c}$ \\[1mm]
$\ivalset{\cup}{\varphi_1\lor\varphi_2} = \ivalset{\cup}{\varphi_1}  \cup  \ivalset{\cup}{\varphi_2}$  \\[1mm]
$\ivalset{\cup}{\varphi_1\land\varphi_2} =\ivalset{\cup}{\varphi_1} \cap \ivalset{\cup}{\varphi_2}$\\[1mm]
$\ivalset{\cup}{\lF_{[a,b]}\varphi} = \big(\bigcup_{I\in\ivalset{}{\varphi}}I\ominus [a,b]\big) \cup [T-b,T)$\\[1mm]
$\ \ivalset{\cup}{\lG_{[a,b]}\varphi} = \big(\bigcup_{I\in(\ivalset{}{\varphi})^{c}}I\ominus [a,b]\big)^{c}\cup [T-a,T)\ $\\[1mm]
$\ \ivalset{\cup}{\varphi\lU_{[a,b]}\psi} = \bigcup_{I_{\varphi}\in\ivalset{}{\varphi}}\bigcup_{I_{\psi}\in\ivalset{}{\psi}} \Big(\big((I_{\varphi}\cap I_{\psi}) \ominus [a,b] \big) \cap I_{\varphi}\Big) \cup I_\dom,$\\[3mm]
\hspace{2cm} where
$I_\dom = \begin{cases}
[\max(T-b, t),T), &\text{if $\exists t$ s.t. $[t,T) \in \ivalset{}{\varphi}$}\\
\emptyset,  &\text{otherwise}
\end{cases}$
\\[4mm]
\hline
\end{tabular}
\vspace{-3mm}
\end{table}
While the table presents the computation of $\ivalset{\cup}{\varphi}$, we can obtain $\ivalset{}{\varphi}$ by simply partitioning $\ivalset{\cup}{\varphi}$ into maximal disjoint intervals.

For an illustration, we consider the example from the introduction and compute $\ivalsetu{}{\varphi_2}$ where $u_1$ is the first positive prefix, $\varphi_2 = p \lor \lF_{[0,1]}q$, and $\dom=[0,6)$. 
First, we have $\ivalsetu{}{p} = \{[0,2),[3,6)\}$ and $\ivalsetu{}{q} = \{[0,1),[2,4)\}$.
Now, we can compute $\ivalsetu{}{\lF_{[0,1]}q} = \{[0,4),[5,6)\}$ and then $\ivalsetu{}{p \lor \lF_{[0,1]}q} = \{[0,6)\}$. Now, we formally prove Lemma~\ref{lem:corr_maler} that proves the correctness of our construction of $\ivalset{}{\varphi}$ given a prefix $\prefix{x}$.

\begin{proof}[Proof of Lemma~\ref{lem:corr_maler}]
	We prove both directions together by induction on the structure of the formula $\varphi$.
	
	For the base case, one can check that for all $t \in [0, T)$, $t \in \ivalset{}{p}$ if and only if $t \in I$ for some $I\in \ivalset{}{\varphi}$ by construction.
	The proof for the $\ neg$ operator and the boolean connectives $\land$ and $\lor$ follow from the correctness of the construction in the work of~\cite{maler}. Here, we provide the proof for the $\lF_{[a,b]}$ operator. The proofs for the $\lU_{[a,b]}$ and $\lG_{[a,b]}$ can be obtained similarly. 
	
	Let $\varphi = \lF_{[a,b]} \psi$. To show the forward direction, let $t \in I$ for some $I \in \ivalset{}{\varphi}$. We have to prove that, $(\prefix{x}, t) \fmodels  \lF_{[a,b]} \psi$. In particular, $t \in \ivalset{\cup}{\varphi}$ by definition, i.e., $t \in \big(\bigcup_{I\in\ivalset{}{\psi}}I\ominus [a,b]\big) \cup [T-b,T)$. There are two cases: (i)  $t \in [T-b,T)$: in this case, $t+b \geq T$ and by definition of $\fmodels$, $(\prefix{x},t) \fmodels \varphi$, or (ii) $t \in \big(\bigcup_{I\in\ivalset{}{\psi}}I\ominus [a,b]\big)$: Fix the interval $I' = [t_1, t_2) \in \ivalset{}{\psi}$ such that, $t \in (I' \ominus [a,b])$. By induction hypothesis, for all $t' \in I'$, $(\prefix{x},t') \fmodels \psi$. Now, $t < t_2 -a \implies t+a <t_2$ and $t \geq t_1 -b \implies t+b \geq t_1$. Hence, $I' = [t_1,t_2) \supset [t+a, t+b]$. Hence, $\exists t' \in [t+a,t+b]$ such that, $(\prefix{x},t') \fmodels \psi$ and henceforth, $(\prefix{x},t) \fmodels \varphi$.
	
	For the backward direction, we assume that, $(\prefix{x}, t) \fmodels  \lF_{[a,b]} \psi$ and prove that, $t \in I$ for some $I \in \ivalset{}{\varphi}$. In particular, we show that $t \in \ivalset{\cup}{\varphi} =  \big(\bigcup_{I\in\ivalset{}{\psi}}I\ominus [a,b]\big) \cup [T-b,T)$ and the rest of the argument follows from the fact that, $\ivalset{}{\varphi}$ is obtained by taking the maximal disjoint intervals of $\ivalset{\cup}{\varphi}$. Now, by definition of $\fmodels$, there are two possibilities: (i) $t+b \geq T$: then, $t \in [T-b, T)$ and hence, $t \in \ivalset{\cup}{\varphi}$, or (ii) $\exists t' \in [t+a, t+b]$ such that, $(\prefix{x},t') \fmodels \psi$. Now, by induction hypothesis, $t' \in I$ for some $I \in \ivalset{}{\psi}$. Let $I = [t_1,t_2)$. Now, $t_2-a > t'-a\geq t$ and $t_1 -b \leq t'-b \leq t$. This implies that, $t \in [t_1 -b, t_2-a) = (I \ominus [a,b])$ which proves that, $t \in \ivalset{\cup}{\varphi}$. \qed
\end{proof}
%\RIT{proof for until might get very complicated}

%The correctness of the procedure can be established by the following lemma:

%The proof of Lemma~\ref{lem:corr_maler} mostly follows from the correctness of the algorithm by Maler et al.~\cite{maler}, except the temporal operators which we present in the appendix.\RAJ{mention appendix here}

In the monitoring algorithm, the number of maximal intervals required in $\ivalset{}{\varphi}$ is upper-bounded by $\mathcal{M}=\mu|\varphi|$, where $\mu = \max(\{|\ivalset{}{p}|~|~ p \in \prop\})$, as also observed by Maler and Nickovic~\cite{maler}.
The computation of this bound can also be done inductively on the structure of $\varphi$.

Now, in the subformula $\semcons$, we symbolically encode the set $\ivalset{}{\varphi}$ of our prospective $\MTL$ formula $\varphi$.
To this end, we introduce variables $t^l_{i,m,s}$ and $t^r_{i,m,s}$ where $i\in \{1,\dots,n\}$, $m\in\{1,\dots,\mathcal{M}\}$, and $s\in \{1,\dots,|\sample|\}$, $s$ being an identifier for the $s^{th}$ prefix $\prefix{x}^{s}$ in $\sample$.
The variables $t^l_{i,m,s}$ and $t^r_{i,m,s}$ encode the $m^{th}$ interval of $\ivalsetx{}{\varphi[i]}$ for the subformula $\varphi[i]$.
In other words, $t^l_{i,m,s}=t_1$ and $t^r_{i,m,s}=t_2$ if and only if $[t_1,t_2)$ is the $m^{th}$ interval of $\ivalsetx{}{\varphi[i]}$.

Now, to ensure that the variables $t^l_{i,m,s}$ and $t^r_{i,m,s}$ have their desired meaning, we introduce constraints for each operator based on the relations defined in Table~\ref{tab:monitoring-algo}.
We now present these constraints for the different $\MTL$ operators.

For the $\neg$ operator, we have the following constraints:
\begin{align*}
	\bigwedge_{\substack{{1\leq i\leq n}\\{1\leq j < i}}} x_{i,\neg} \wedge l_{i,j} \rightarrow \big[\bigwedge_{1\leq s\leq|\sample|}\mathit{comp}_{s}(i,j)\big], 
\end{align*}
where, for every $\prefix{x}^s$ in $\sample$, $\mathit{comp}_{s}(i,j)$ encodes that $\ivalsetx{\cup}{\varphi[i]}$ is the complement of $\ivalsetx{\cup}{\varphi[j]}$ .
We construct $\mathit{comp}_s(i,j)$ as follows:
\begin{align}
	\mathtt{ite}(t^l_{j,1,s} = 0,&\label{eq:neg-condition} \\
	&\bigwedge_{1\leq m\leq \mathcal{M}-1} t^l_{i,m,s} = t^r_{j,m,s} \wedge t^r_{i,m,s} = t^l_{j,m+1,s},\label{eq:neg-case1}\\
	&t^l_{i,1,s} = 0 \wedge t^r_{i,1,s} = t^l_{j,1,s} \wedge \label{eq:neg-case2}\\
	&\bigwedge_{1\leq m\leq \mathcal{M}-1} t^l_{i,m+1,s} = t^r_{j,m,s} \wedge t^r_{i,m+1,s} = t^l_{j,m+1,s}),\nonumber
\end{align}
%where $\mathtt{ite}$ (short for “if-then-else”) is syntactic sugar for a conditional evaluation of formulas, which is a standard construct for SMT solvers.
where $\mathtt{ite}$ is a syntactic sugar for the ``if-then-else'' construct over LRA formulas, which is standard in many SMT solvers.
Here, Condition~\ref{eq:neg-condition} checks whether the left bound of the first interval of $\ivalsetx{}{\varphi[j]}$, encoded by $t^l_{j,1,s}$, is $0$. 
If that holds, as specified by Constraint~\ref{eq:neg-case1}, the left bound of the first interval of $\ivalsetx{}{\varphi[i]}$, encoded by $t^l_{1,i,s}$, will be the right bound of the first interval of $\ivalsetx{}{\varphi[j]}$, encoded $t^r_{1,j,s}$ and so on.
If Condition~\ref{eq:neg-condition} does not hold, as specified by Constraint~\ref{eq:neg-case2}, the left bound of the first interval of $\ivalsetx{}{\varphi[i]}$
%\RIT{we can remove this}, encoded by $t^l_{1,i,s}$, 
will start with 0, and so on. 

As an example, for a prefix $\prefix{x}^s$ and $\dom=[0,7)$, let $\ivalsetx{}{\varphi[j]}= \{[0,4), [6,7)\}$. Then, Constraint~\ref{eq:neg-case1} ensures that $\ivalsetx{}{\varphi[i]} = \{[4,6)\}$\footnote{ $|\ivalsetx{}{\varphi[i]}|$ may differ for different subformulas $\varphi[i]$; we address this at the end of this section.}.
Conversely, if $\ivalsetx{}{\varphi[j]}= \{[1,4), [6,7)\}$, then Constraints~\ref{eq:neg-case2} ensures that $\ivalsetx{}{\varphi[i]} = \{[0,1),[4,6)\}$.
%\RIT{s in the prefix?}
%\RIT{another example in one line for the other case?}

For the $\lor$ operator, we have the following constraint:
\begin{align*}
	\bigwedge_{\substack{{1\leq i\leq n}\\{1\leq j,j' < i}}} x_{i,\lor} \wedge l_{i,j} \wedge r_{i,j'} \rightarrow \big[\bigwedge_{1\leq s\leq|\sample|}\mathit{union}_{s}(i,j,j')\big], 
\end{align*}
where, for every $\prefix{x}^s$ in $\sample$, $\mathit{union}_{s}(i,j,j')$ encodes that $\ivalsetx{}{\varphi[i]}$ consists of the maximal disjoint intervals obtained from the union of the intervals in $\ivalsetx{}{\varphi[j]}$ and $\ivalsetx{}{\varphi[j']}$.
We construct $\mathit{union}_{s}(i,j,j')$ as follows:
\begin{align}
	&\bigwedge_{\sigma \in \left[l,r\right]}\bigwedge_{1\leq m\leq \mathcal{M}} \left(\bigvee_{1\leq m' \leq \mathcal{M}}(t^{\sigma}_{i,m,s} =   t^{\sigma}_{j,m',s}) \lor \bigvee_{1\leq m' \leq \mathcal{M}} (t^{\sigma}_{i,m,s} =   t^{\sigma}_{j',m',s})\right) \land\label{eq:union-case1-main}\\
	& \bigwedge_{\sigma \in \left[l,r\right]}\bigwedge_{1 \leq m \leq \mathcal{M}} \left(\bigvee_{1\leq m' \leq \mathcal{M}}(t^\sigma_{i,m,s} =  t^\sigma_{j,m',s}) \iff \bigwedge_{1\leq m'' \leq \mathcal{M}}(t^\sigma_{j,m',s} \not \in I_{j',m'',s})\right)\land\label{eq:union-case2-main}\\
	& \bigwedge_{\sigma \in \left[l,r\right]}\bigwedge_{1 \leq m \leq \mathcal{M}} \left(\bigvee_{1\leq m' \leq \mathcal{M}}(t^\sigma_{i,m,s} =  t^\sigma_{j',m',s}) \iff \bigwedge_{1\leq m'' \leq \mathcal{M}}(t^\sigma_{j',m',s} \not \in  I_{j,m'',s})\right),\label{eq:union-case3-main}
\end{align}
where $I_{k,m,s}$ denotes the interval encoded by bounds $t^l_{k,m,s}$ and $t^r_{k,m,s}$\footnote{In LRA, $t \not\in [t_1,t_2)$ can be encoded as $t< t_1 \vee t\ge t_2$.}.
Here, Constraint~\ref{eq:union-case1-main} states that the left (resp., right) bound of each interval of $\ivalsetx{}{\varphi[i]}$, encoded by $t^l_{i,m,s}$ (resp., $t^r_{i,m,s}$) corresponds to one of the left (resp., right) bounds of the intervals in $\ivalsetx{}{\varphi[j]}$ or in $\ivalsetx{}{\varphi[j']}$.
Then, Constraint~\ref{eq:union-case2-main} states that for each interval $I$ in $\ivalsetx{}{\varphi[j]}$, the left (resp., right) bound of $I$ should appear as the left (resp., right) bound of some interval in $\ivalsetx{}{\varphi[i]}$ if and only if the left (resp., right) bound of $I$ is not included in any of the intervals in $\ivalsetx{}{\varphi[j']}$.
Constraint~\ref{eq:union-case3-main} mimics the statement made by Constraint~\ref{eq:union-case2-main} but for the bounds of the intervals in $\ivalsetx{}{\varphi[j']}$.

For an illustration, assume that $\ivalsetx{}{\varphi[j]} = \{[1,4),[6,7)\}$ and $\ivalsetx{}{\varphi[j']} = \{[3,5),[6,7)\}$ for a prefix $\prefix{x}^s$ and $T=7$.
Now, if $\varphi[i]=\varphi[j]\vee\varphi[j']$, then $\ivalsetx{}{\varphi[i]}=\{[1,5),[6,7)\}$ based on the relation for $\vee$-operator in Table~\ref{tab:monitoring-algo}.
Observe that all the bounds of the intervals in $\ivalsetx{}{\varphi[i]}$, i.e., 1, 5, 6, and 7, are present as the bounds of the intervals in either $\ivalsetx{}{\varphi[j]}$ or $\ivalsetx{}{\varphi[j']}$.
This fact is in accordance with Constraint~\ref{eq:union-case1-main}. 
Also, the right bound of $[1,4)$ in $\ivalsetx{}{\varphi[j]}$ does not appear as a bound of any intervals in $\ivalsetx{}{\varphi[i]}$, as it is included in an interval in $\ivalsetx{}{\varphi[j']}$, i.e., $4 \in [3,5)$.
This is in accordance with Constraint~\ref{eq:union-case2-main}.

Next, for the $\lF_I$-operator where $I$ is encoded using $a_i$ and $b_i$, we have the following constraint:
\begin{align*}
	\bigwedge_{\substack{{1\leq i\leq n}\\{1\leq j< i}}} x_{i,\lF_I} \wedge l_{i,j} \rightarrow \big[\bigwedge_{1\leq s\leq|\sample|} \mathit{union}'_{s}(i,k,k) \wedge \ominus^{[a_i,b_i]}_s(k,j) \big].
\end{align*}
based on the relation for the $\lF_{[a,b]}$ operator in Table~\ref{tab:monitoring-algo}.
We here rely on an intermediate set of intervals $\tilde{\I}_k$ encoded using some auxiliary variables $\tilde{t}^l_{k,m,s}$ and $\tilde{t}^r_{k,m,s}$ where $m\in\{1,\ldots,\mathcal{M}\}$ and $s\in\{1,\dots,|\sample|\}$.
Also, we use the formula $\ominus^{[a_i,b_i]}_s(k,j)$ to encode that the intervals in $\tilde{\I}_k$ can be obtained by performing $I\ominus[a,b]$ to each interval $I$ in $\ivalsetx{}{\varphi[j]}$, where $a_i=a$ and $b_i=b$.
Finally, the formula $\mathit{union}'(i,k,k)$ encodes that $\ivalsetx{}{\varphi[i]}$ consists of the maximal disjoint intervals obtained from the union of the intervals in $\tilde{\I}_k$ and $\{[T-b,T)\}$.

The construction of $\mathit{union}'(i,k,k)$ is similar to that of $\mathit{union}(i,j,j')$ in that the constraints involved are similar to Constraints~\ref{eq:union-case1-main} to~\ref{eq:union-case3-main}.
For $\ominus^{[a_i,b_i]}_s(k,j)$, we have the following constraint:
\begin{align}\label{eq:minus}
	\bigwedge_{1 \leq m \leq \mathcal{M}-1} \big[\tilde{t}^l_{k,m,s} = \max\{0,\left(t^l_{j,m,s} -b_i\right)\}\ \land \tilde{t}^r_{k,m,s} = \max\{0,\left(t^r_{j,m,s} -a_i\right)\}\big]
\end{align}

As an example, consider $\ivalsetx{}{\varphi[j]}=\{[1,4),[6,7)\}$
for a prefix $\prefix{x}^s$ and $T=7$.
Now, if $\varphi[i] = \lF_{[1,4]}\varphi[j]$, then first we have $\tilde{\I}_k = \{[0,3),[2,6)\}$ based on Constraint~\ref{eq:minus}
\footnote{While the intervals in $\tilde{\I}_k$ may not be disjoint, $\mathit{union}'(i,k,k)$ ensures that $\ivalsetx{}{\varphi[i]}$ consists of only maximal disjoint intervals.}.
Next, we have $\ivalsetx{}{\varphi[i]}=\{[0,7)\}$ which consists of the maximal disjoint intervals from $\tilde{\I}^{}_k \cup \{[T-4,T)\} = \{[0,3),[2,6),[3,7)\}$ using $\mathit{union}'(i,k,k)$.
%According to Constraint~\ref{eq:minus}, while $\tilde{\I}^{\cup}_k$ (i.e., the union of all intervals in $\tilde{\I}_k$) contains all the time points where $\lF_{[a,b]}$ holds, $\tilde{\I}_k$ might not only contain maximal disjoint intervals.

For the $\lU_I$ operator, we have the following constraint:
\begin{align*}
	\bigwedge_{\substack{{1\leq i\leq n}\\{1\leq j,j' < i}}} x_{i,\lU_I} \wedge l_{i,j} \wedge r_{i,j'} \rightarrow \big[\bigwedge_{1\leq s\leq|\sample|}\mathit{intersection}_{s}(k_1,j,j') 
    \wedge \ominus^{[a_i,b_i]}_s(k_2,k_1) \\\wedge  \mathit{cond{-}int}_s(k_3,k_2,j) \wedge \mathit{union}_s(i, k_3, k_3)                             
 \big]
\end{align*}
Here, we introduce three intermediate set of intervals $\tilde{\I}_{k_1}$, $\tilde{\I}_{k_2}$ and $\tilde{\I}_{k_3}$ encoded using auxiliary variables $\tilde{t}^l_{k_i,m,s}$ and $\tilde{t}^r_{k_i,m,s}$ where $i \in \{1,2,3\}$, $m\in\{1,\ldots,\mathcal{M}\}$ and $s\in\{1,\dots,|\sample|\}$. 
Similar to the constraints for the $\lor$ operator, we denote an interval in $\tilde{\I}_{k_i}$ as $I_{k_i,m,s}$ where, $I_{k_i,m,s} = [\tilde{t}^l_{k_i,m,s}, \tilde{t}^r_{k_i,m,s})$. Now, $\mathit{intersection}_{s}(k_1,j,j')$ encodes that $\tilde{\I}_{k_1}$ consists of the maximal disjoint intervals obtained from the intersection of the intervals in $\ivalsetx{}{\varphi[j]}$ and $\ivalsetx{}{\varphi[j']}$. Note that the intersection can be achieved using the $\mathit{union}_s$ and the $\mathit{comp}_s$ operators using De Morgan's law, i.e., $A \cap B = (A^c \cup B^c)^c$. Then, $\ominus^{[a_i,b_i]}_s(k_2,k_1)$ denotes that the intervals in $\tilde{\I}(k_2)$ can be obtained by performing $I \ominus [a,b]$ to each interval in $\tilde{\I}_{k_1}$ using constraint~\ref{eq:minus}. Next, the operator $\mathit{cond-int}_s(k_3,k_2,j)$ denotes that the $m^{th}$ interval in $\tilde{\I}(k_3)$ ($I_{k_3,m,s}$) is obtained by taking the intersection of the $m^{th}$ interval in $\tilde{\I}(k_2)$ ($I_{k_2,m,s}$) and the $m'^{th}$ interval in $\ivalsetx{}{\varphi[j]}$ ($I_{j,m',s}$) such that, $I_{k_1,m,s}$ ($I_{k_2,m,s} = I_{k_1,m,s} \ominus [a,b]$, by construction) is a subset of $I_{j,m',s}$. This can be achieved by encoding $\mathit{cond-int}_s(k_3,k_2,j)$ as the following constraint:

\begin{align*}
	\bigwedge_{1\leq m\leq \mathcal{M}} \bigwedge_{1\leq m'\leq \mathcal{M}} (I_{k_1,m,s} \subseteq I_{j,m',s}) \rightarrow I_{k_3,m,s} = I_{k_2,m,s} \cap I_{j,m',s}
\end{align*}

Note that the subset check and the intersection of two intervals both allow simple encodings in LRA. Finally, the formula $\mathit{union}_s(i,k_3, k_3)$ encodes that $\ivalsetx{}{\varphi[i]}$ consists of the maximal disjoint intervals obtained from the union of the intervals in $\tilde{\I}_{k_3}$.

For an illustration, assume that $\ivalsetx{}{\varphi[j]} = \{[1,3),[5,8)\}$ and $\ivalsetx{}{\varphi[j']} = \{[4,6),[7,9)\}$ for a prefix $\prefix{x}^s$ and $T=9$.
Now, let $\varphi[i]=\varphi[j]\lU_{[0,3]}\varphi[j']$. Then, $\ivalsetx{}{\varphi[i]} = \{[5,8)\}$ using the computation in Table~\ref{tab:monitoring-algo}.

Note that, following the constraint, $\tilde{\I}_{k_1} = \{[5,6), [7,8)\}$ after taking the intersection of $\ivalsetx{}{\varphi[j]}$ and $\ivalsetx{}{\varphi[j']}$. Then, the Minkowski minus results into the set of intervals $\tilde{\I}_{k_2} = \{[2,6), [4,8)\}$ with $a =0$ and $b=3$. The conditional intersection of $\tilde{\I}_{k_2}$ and $\ivalsetx{}{\varphi[j}$ produces the set of intervals $\tilde{\I}_{k_3} = \{[5, 6), [5,8)\}$. Note that this is because both the intervals in $\tilde{\I}_{k_1}$ are subsets of the interval $[5,8)$ in $\ivalsetx{}{\varphi[j]}$ and not of $[1,3)$ and we intersect the intervals in $\tilde{\I}_{k_2}$ with only $[5,8)$. Finally the operator $\mathit{union}_s$ on $\tilde{\I}_{k_3}$ results in $\ivalsetx{}{\varphi[i]}$ to be $\{[5,8)\}$ that complies with the actual semantics of the $\lU_I$ operator. It can be also checked that taking a normal intersection instead of the conditional one would have wrongly resulted in $\ivalsetx{}[\varphi[i]]$ to be $\{[2,3), [5,8)\}$ that depicts the intricacy in computing the satisfaction intervals for $\lU_I$ as shown in Figure~3(a) in~\cite{maler}.

For the $\land$-operator, we have the following constraint:
\begin{align*}
	\bigwedge_{\substack{{1\leq i\leq n}\\{1\leq j,j' < i}}} x_{i,\land} \wedge l_{i,j} \wedge r_{i,j'} \rightarrow \big[\bigwedge_{1\leq s\leq|\sample|}\mathit{intersection}_{s}(i,j,j')\big], 
\end{align*}
This encodes the relation for $\land$ operator as described in Table~\ref{tab:monitoring-algo}, i.e., encoding the fact that the set $\ivalsetx{\cup}{\varphi[i]}$ contains maximal disjoint intervals of \emph{intersection} of $\ivalsetx{\cup}{\varphi[j]}$ and $\ivalsetx{\cup}{\varphi[j']}$.

For the $\lG_I$ operator where $I$ is encoded using $a_i$, $b_i$ we have the following constraint:

\begin{align*}
	\bigwedge_{\substack{{1\leq i\leq n}\\{1\leq j< i}}} x_{i,\lG_I} \wedge l_{i,j} \rightarrow \big[\bigwedge_{1\leq s\leq|\sample|} \mathit{union}''_{s}(i,k',k') \wedge \mathit{comp}_s(k',k) \wedge \ominus^{[a_i,b_i]}_s(k,j) \big].
\end{align*}
based on the relation for the $\lG_{[a,b]}$ operator in Table~\ref{tab:monitoring-algo}.
Similar to the encoding of $\lF_{I}$ operator, we rely on an intermediate set of intervals $\tilde{\I}_k$ and $\tilde{\I}_{k'}$ encoded using some auxiliary variables.
Also, we use the formula $\ominus^{[a_i,b_i]}_s(k,j)$ to encode that the intervals in $\tilde{\I}_k$ can be obtained by performing $I\ominus[a,b]$ to each interval $I$ in $\ivalsetx{}{\varphi[j]}$, where $a_i=a$ and $b_i=b$. Then $\mathit{comp}_s(k',k)$ encodes that $\tilde{\I}_{k'}$ is the complement of $\tilde{\I}_{k}$.
Finally, the formula $\mathit{union}''(i,k',k')$ encodes that $\ivalsetx{}{\varphi[i]}$ consists of the maximal disjoint intervals obtained by taking the union of the complement of $\ivalsetx{\cup}{\varphi[j]}$ and $\{[T-a,T)\}$.

Similar to $\mathit{union'}$ in the semantic constraints for $\lF_I$ operator, the construction of $\mathit{union}''(i,k,k)$ is similar to that of $\mathit{union}(i,j,j')$ in that the constraints involved are similar to Constraints~\ref{eq:union-case1-main} to~\ref{eq:union-case3-main}.

%For the $\lG_I$ and the $\wedge$ operator, we encode the relations described in Table~\ref{tab:monitoring-algo} by reusing the constraints from the formulas $\mathit{comp}_s(i,j)$, $\mathit{union}_s(i,j,j')$, and $\ominus^{[a_i,b_i]}_s(k,j)$.
%We present the exact constraints in Appendix~\ref{sec: app_semantics}.
We now assert the correctness of the formulas encoding the set operations as follows: 
\begin{lemma}\label{lem:set-operation-correctness}
The formulas $\mathit{comp}_s(i,j)$, $\mathit{union}_s(i,j,j')$, $\ominus^{[a_i,b_i]}_s(k,j)$, $\mathit{intersection}_s(i,j,j')$and $\mathit{cond{-}int}_s(i,j,j')$ correctly encode the complement, union, $\ominus$, intersection and conditional intersection operations on a set of intervals, resp.
\end{lemma}
%The proof of the lemma is presented in Appendix~\ref{sec:set-correct}.
\begin{proof}[Proof of Lemma~\ref{lem:set-operation-correctness}] Here, we provide the proof of the correctness of the construction of each formula mentioned in Lemma~\ref{lem:set-operation-correctness}.

\begin{itemize}
    \item \begin{claim}[Correctness of $\mathit{union}_s$]
		Let $\inter$ be a satisfying interpretation of $\mathit{union}_s(i,j,j')$. Then, the set $\I_i = \{[\inter(t^l_{i,1,s}), \inter(t^r_{i,1,s})),\dots,[\inter(t^l_{i,m,s}), \inter(t^r_{i,m,s}))\}$ consists of the maximal disjoint intervals of the union of $\I_j = \{[\inter(t^l_{j,1,s}), \inter(t^r_{j,1,s})),\dots, [\inter(t^l_{j,m,s}), \inter(t^r_{j,m,s}))\}$ and $\I_{j'} = \{[\inter(t^l_{j',1,s}), \inter(t^r_{j',1,s})),\dots, [\inter(t^l_{j',m,s}), \inter(t^r_{j',m,s}))\}$.
  \end{claim}
	\begin{proof}
		For simplicity of the proof, we name $\inter(t^\sigma_{\kappa,m})$ as $\tau^\sigma_{\kappa,m}$ for $\sigma\in \{l,r\}$ and $\kappa\in\{i,j,j'\}$, and $[\tau^l_{\kappa,m},\tau^r_{\kappa,m})$ as $\Gamma_{\kappa,m}$ for $\kappa\in\{i,j,j'\}$.
		Note that we drop the identifier $s$ representing the prefix since the prefix is fixed throughout the proof.
		
		For the forward direction, we show that any time point $t\in \Gamma_{i,m}$ belongs to some $\Gamma_{j,m'}\in\I_{j}$ or some $\Gamma_{j',m''}\in\I_{j'}$.
		Towards contradiction, we assume that $t\not\in\Gamma_{j,m'}$ for any $\Gamma_{j,m'}\in\I_{j}$ and $t\not\in\Gamma_{j',m''}$ for any  $\Gamma_{j',m''}\in\I_{j'}$.
		Now, based on Constraint~\ref{eq:union-case1-main}, both $\tau^l_{i,m}$ and $\tau^r_{i,m}$ appear in some intervals in $\I_j$ and $\I_{j'}$ as left and right bound, respectively.
		We consider two cases based on where $\tau^l_{i,m}$ and $\tau^r_{i,m}$ appear.
		First, $\tau^l_{i,m}$ and $\tau^r_{i,m}$ both appears, w.l.o.g, in $\I_j$.
		Now, let $\Gamma_{j,m_1}$ and $\Gamma_{j,m_1+1}$ be such that $\tau^r_{j,m_1}\leq t < \tau^l_{j,m_1+1}$. 
		Intuitively, this means that $t$ lies in between (and is adjacent to) the intervals $\Gamma_{j,m_1}$ and $\Gamma_{j,m_1+1}$.
		Note that both $\tau^r_{j,m_1}$ and $\tau^l_{j,m_1+1}$ is not included in $\I_i$ since $\I_i$ consists of maximal disjoint intervals and $[\tau^r_{j,m_1},\tau^l_{j,m_1+1}] \subset \Gamma_{i,m}$.
		Now, based on Constraint~\ref{eq:union-case2-main}, $\tau^r_{j,m_1}$ and $\tau^l_{j,m_1+1}$ are included in some intervals in $\I_{j'}$. Note that if they are included in the same interval, then that interval also contains $t$ raising the contradiction to our assumption that $t\not\in\Gamma_{j',m''}$ for any  $\Gamma_{j',m''}\in\I_{j'}$. Then $\tau^r_{j,m_1}$ and $\tau^l_{j,m_1+1}$ are not included in the same interval in $\I_{j'}$. Then, there exists $\Gamma_{j',m_2} \in \I_{j'}$ and $\Gamma_{j',m_2+1}\in \I_{j'}$ such that, 
		\[
		\tau^r_{j,m_1}< \tau^r_{j',m_2} \leq t < \tau^l_{j',m_2+1}< \tau^l_{j,m_1+1}
		\]
		Now note that, $\tau^r_{j',m_2}$ and $\tau^l_{j',m_2+1}$ both are not included in any of the intervals in $\I_j$. Now, based on Constraint~\ref{eq:union-case3-main}, both appear in $\I_i$. But that raises the contradiction to our assumption that $t \in \Gamma_{i,m}$.

		For the other direction, we show that any time point, w.l.o.g, $t\in \Gamma_{j,m}$ belongs to some $\Gamma_{i,m'}\in\I_{j}$.
		For this, there can be three cases based on whether the bounds of
		$\Gamma_{j,m}$ appear as bounds in some interval $\Gamma_{i,m'}\in \I_i$ or not.
		
		First, assume that both $\tau^l_{j,m}$ and $\tau^r_{j,m}$ appear as bounds $\tau^{l}_{i,m_1}$ and $\tau^{r}_{i,m_2}$ in $\I_k$ as stated by Constraint~\ref{eq:union-case1-main}.
		We now claim that $m_1=m_2$ meaning that $\tau^l_{i,m_1}$ and $\tau^r_{i,m_2}$ are bounds of the same intervals.
		Towards contradiction, let $m_1+1\leq m_2$.
		Then, $\tau^r_{i,m_1}$ belongs to the interval $\Gamma_{j,m}$, and based on Constraint~\ref{eq:union-case2-main}, and cannot be one of the bounds of $\Gamma_{i,m_1}$.
		Then, we have $\tau^{l}_{j,m}=\tau^{l}_{i,m_1}\leq t < \tau^{r}_{i,m_1}=\tau^{r}_{j,m}$
		
		Second, assume that $\tau^l_{j,m}$ does not appear, while $\tau^r_{j,m}$ appears as bounds in $\I_k$.
		Now, based on Constraint~\ref{eq:union-case2-main}, $\tau^l_{j,m}$ appears in one of the intervals $\Gamma_{j',m'}$ in $\I_{j'}$.
		Also, in that case, $\tau^{l}_{j',m'}$ appears as a left bound in $\I_k$, say $\I_{i,m_1}$.
		We now claim that $\tau^{r}_{i,m_1}>\tau^{r}_{j,m}$.
		Towards contradiction, we assume two cases. 
		In first case, 
		\[
		\tau^{l}_{j',m'}=\tau^{l}_{i,m_1}<\tau^{r}_{i,m_1}<\tau^{l}_{j,m}<\tau^{r}_{j',m}
		\]
		contradicting Constraint~\ref{eq:union-case2-main}.
		In the second case,
		\[
		\tau^{l}_{j',m'}=\tau^{l}_{i,m_1}<\tau^{l}_{j,m}<\tau^{r}_{i,m_1}<\tau^{r}_{j,m}
		\]
		contradicting Constraint~\ref{eq:union-case3-main}.
		From the two cases, we conclude $\tau^{r}_{i,m_1}>\tau^{r}_{j,m}$ and hence, $\tau^{l}_{i,m_1}<\tau^{l}_{j,m}\leq t < \tau^{r}_{j,m} < \tau^{r}_{i,m_1}$.
		The argument in the third case is similar to those in the other two cases and can be seen easily.
	\end{proof}
	
	\item \begin{claim}[Correctness of $\mathit{comp}_s$]
		Let $\inter$ be a satisfying interpretation of $\mathit{comp}_s(i,j)$. Then, the set $\I_i = \{[\inter(t^l_{i,1,s}), \inter(t^r_{i,1,s})),\dots,[\inter(t^l_{i,m,s}), \inter(t^r_{i,m,s}))\}$ consists of the maximal disjoint intervals of the complement of $\I_j = \{[\inter(t^l_{j,1,s}), \inter(t^r_{j,1,s})),\dots, [\inter(t^l_{j,m,s}), \inter(t^r_{j,m,s}))\}$.
	\end{claim}
	\begin{proof}
		We reuse the naming conventions for $\tau^\sigma_{\kappa,m}$ and $\Gamma_{\kappa,m}$ from the last proof.
		For the forward direction, we show that if $t\in \Gamma_{i,m}$ for some $\Gamma_{i,m}\in\I_i$ then $t\not\in\Gamma_{j,m'}$ for any $\Gamma_{j,m'}\in\I_j$.
		First, let $m=1$. 
		Then, if $\tau^{l}_{j,1}=0$, then Condition~\ref{eq:neg-condition} gets triggered and $\tau^{l}_{i,1} = \tau^{r}_{j,1}$ and $\tau^{r}_{i,1} = \tau^{r}_{j,2}$.
		Hence, $\tau^{r}_{j,1}=\tau^{l}_{i,1} \leq t < \tau^{r}_{i,1}=\tau^{l}_{j,2}$.
		Also, if $\tau^{l}_{j,1}\neq 0$, then Condition~\ref{eq:neg-condition} does not get triggered and $\tau^{l}_{i,1} = 0$ and $\tau^{r}_{i,1} = \tau^{l}_{j,1}$.
		Hence, $0=\tau^{l}_{i,1} \leq t < \tau^{r}_{i,1}=\tau^{l}_{j,1}$.
		For $m\neq 1$, the reasoning works similarly.
		
		For the other direction, we show that if $t\in \Gamma_{j,m}$ for some $\Gamma_{j,m}\in\I_j$ then $t\not\in\Gamma_{i,m'}$ for any $\Gamma_{i,m'}\in\I_j$.
		The proof for this direction is almost identical to the proof for the forward direction.
	\end{proof}

	\item \begin{claim}[Correctness of $\ominus^{[a_i,b_i]}_s$]
		Let $\inter$ be a satisfying interpretation of $\ominus^{[a_i,b_i]}_s(k,j)$. Then, the set $\I_i = \{[\inter(t^l_{i,1,s}), \inter(t^r_{i,1,s})),\dots,[\inter(t^l_{i,m,s}), \inter(t^r_{i,m,s}))\}$ consists of the maximal disjoint intervals by applying $I\ominus [a,b]$ to the intervals $I$ of $\I_j = \{[\inter(t^l_{j,1,s}), \inter(t^r_{j,1,s})),\dots, [\inter(t^l_{j,m,s}), \inter(t^r_{j,m,s}))\}$, where $\inter(a_i) = a$ and $\inter(b_i) = b$.
	\end{claim}
	\begin{proof}
		The proof of the above claim follows directly from the construction of the formula $\ominus^{[a_i,b_i]}_s(k,j)$.
	\end{proof}
 \end{itemize}
	
	The correctness of the formulas $\mathit{intersection}_s$ and $\mathit{cond-int}_s$ follow from the correctness of $\mathit{union}_s$ and can be derived using minor modifications.  \qed
	
\end{proof}

It is worth noting that although the number of intervals in $\ivalsetx{}{\varphi[i]}$ for each subformula $\varphi[i]$ is bounded by $\mathcal{M}$, it may not contain the same number of intervals.
For instance, $\ivalsetx{}{p}= \{[0,1),[6,7)\}$ has two intervals, while, assuming $T=7$, $\ivalsetx{}{\neg p} = \{[1,6)\}$ has only one interval.

To circumvent this, we introduce some variables $num_{i,s}$ for $i \in \{1, \dots, n\}$  and $s \in \{1, \dots, |\sample|\}$ to track of the number of intervals in $\ivalsetx{}{\varphi[i]}$ for each subformula $\varphi[i]$ for each prefix $\prefix{x}^s$. 
We now impose $\bigwedge_{1\leq i\leq n, 1\leq m\leq \mathcal{M}} [m > \mathit{num}_{i,s}]\rightarrow [t^l_{i,m,s} = T \land t^r_{i,m,s} = T]$.
This ensures that all the unused variables $t^\sigma_{i,m,s}$ for each Node~$i$ and prefix $\prefix{x}^s$ in $\sample$ are all set to $T$.
We also use the $num_{i,s}$ variables in the constraints for easier computation of $\ivalsetx{}{\varphi[i]}$ for each operator.
We include this in our implementation but omit it here for a simpler presentation.
%Now, for each node~$i$, while calculating the positive intervals, constraints for each operator $\lambda$ does the calculations till $num_{i,s}$ and for all $num_{i,s}<j\leq \mathcal{M}$, we make $I_{i,m,s} = [T,T]$.\RIT{maybe some other bigger number that we track? } 

Finally, to ensure that the prospective formula $\varphi$ is $\gsep$ for $\sample$, we add:
\begin{align*}
	\bigwedge_{\prefix{x}^s \in P} \big[(t^l_{n,1,s} = 0) \land (t^r_{n,1,s} = T) \big] \wedge \bigwedge_{\prefix{x}^s\in N} \big[(t^l_{n,1,s} \neq 0) \lor (t^r_{n,1,s} \neq  T) \big].
\end{align*}
This constraint says that $\ivalsetx{}{\varphi[n]}=\{[0, T)\}$ for all the positive prefixes $\prefix{x}^s$, while $\ivalsetx{}{\varphi[n]}\neq\{[0, T)\}$ for any negative prefixes $\prefix{x}^s$.

%\subsection{The full algorithm.}
%Using the above ideas and the encodings, we give the overview of our full algorithm in this section. We start our search for separating formulas of size 1 and increase the size at each iteration. Precisely, at the $n^{th}$ iteration, we search for 
%separating formulas of size $n$ by constructing the formula $\Phi^n$. Now, if $\Phi_n$ is satisfiable, we take the satisfying assignment $v$ and construct the separating $\MTL$ formula $\varphi_v$ and output it as the smallest size separator. If $\Phi_n$ is not satisfiable, we move to the next iteration and repeat our search process. In practice, the specification cannot be large enough to be efficiently monitorable. Hence, we assume that we terminate the search process once a realistic size bound is reached and no separating formula could be found. The algorithm is outlined in Algorithm~\ref{alg:main}.

The correctness of our algorithm follows from the correctness of the inductive computation of $\ivalset{}{\varphi}$ in Lemma~\ref{lem:corr_maler} and its encoding using the formulas described in Lemma~\ref{lem:set-operation-correctness}.
We state the correctness result formally as follows:
\begin{theorem}[Correctness]\label{thm:correctness}
Given a sample $\sample$ and a future-reach bound $\bound$, Algorithm~\ref{alg:main} terminates and outputs a minimal $\MTL$ formula $\varphi$ such that $\varphi$ is globally separating for $\sample$ and $\fr(\varphi)\leq K$, if such a formula exists.
\end{theorem}
\begin{proof}
The termination of Algorithm~\ref{alg:main} is guaranteed by the decision procedure of checking whether $\sample$ is $\gksepable$ (Section~\ref{sec:existence}).
The minimality of the synthesized formula is due to the iterative search of formulas of increasing size and the correct encoding of $\Phi^{n}_{\sample,K}$.
The correctness of $\Phi^{n}_{\sample,K}$ follows from the correctness of the encoding of set operations described in Lemma~\ref{lem:set-operation-correctness} and the correctness of computation of the sets $\ivalset{}{\varphi}$ using Lemma~\ref{lem:corr_maler}. \qed
\end{proof}

Our synthesis algorithm solves the optimization problem $\syn$ by constructing formulas in LRA. 
We now analyze the computational hardness of $\syn$ and, thus, consider its corresponding decision problem $\synd$: given a sample $\sample$, a future-reach bound $\bound$ and size bound $B$ (in unary), does there exist an $\MTL$ formula $\varphi$ such that $\varphi$ is $\gsep$ for $\sample$, $\fr(\varphi) \le \bound$, and $|\varphi| \le B$. 
Following our algorithm, we can encode the $\synd$ problem in an LRA formula $\Phi= \bigvee_{n \le B}\Phi^n_{\sample,\bound}$, where $\Phi^n_{\sample,\bound}$ is as described in Algorithm~\ref{alg:main}. 
One can check that the size of $\Phi$ is $\mathcal{O}(|\sample||K|B^3\mathcal{M}^3)$. 
Now, the fact that the satisfiability of an LRA formula is $\NP$-complete~\cite{Barrett2018} proves the following:
\begin{theorem}
	$\synd$ is in $\NP$.
\end{theorem}

\begin{remark} \label{thm:lowerbound}
While the exact complexity lower bound for $\synd$ is unknown, we conjecture that $\synd$ is $\NP$-hard.
Our hypothesis stems from the fact that the problem is already $\NP$-hard for simple fragments of $\LTL$~\cite{FijalkowL21}. Note that the hardness result does not directly extend to MTL: the complexity might be either lower or higher since the logic is a priori more expressive. We leave the hardness result for full $\MTL$ as an open problem.
\end{remark}

%% file: experiments.tex
\label{sec:experiments}
In this section, we answer the following research questions to assess the performance of our algorithm for synthesizing $\MTL$ formulas.
\begin{description}
\item[RQ1:] Can our algorithm synthesize concise formulas with small future-reach?
\item[RQ2:] How does lowering the future-reach bound affect the size of the formulas?
\item[RQ3:] How does our algorithm scale for different sample sizes?
\end{description}

To answer the research questions above, we have implemented a prototype of our algorithm in Python 3 using Z3~\cite{MouraB08} as the SMT solver in a tool named \tool{} (synThesizing Efficiently monitorAble mtL). 
To our knowledge, \tool{} is the only tool for synthesizing minimal $\MTL$ formulas for monitoring purposes (see related works).
In \tool{}, we implement a heuristic on top of Algorithm~\ref{alg:main}.
We initially set the maximum number of intervals $\mathcal{M}$ in sets $\ivalset{}{\varphi[i]}$ to be $\mu+2$ where $\mu = \max(\{|\ivalset{}{p}|~|~ p \in \prop\})$.
We iteratively increase the value of $\mathcal{M}$ until we find a solution. 
To ensure that the synthesized $\MTL$ formulas are correct, we implement a verifier based on the inductive computation of $\ivalset{}{\varphi}$ mentioned in Table~\ref{tab:monitoring-algo}.
The heuristic improves the runtime of \tool{} significantly since most $\gsep$ formulas $\varphi$ never require the worst-case upper bound\footnote{The operators $\lF_I$, $\lG_I$, $\land$, and $\neg$ increase the number of required intervals by at most one. Only the $\lor$ operator can double it in the worst-case.} of $\mathcal{M}=\mu|\varphi|$.

As typically done in the literature of synthesizing formulas~\cite{flie,ArifLERCT20,scarlet}, we evaluate \tool{} on benchmarks generated synthetically from $\MTL$ formulas.
To obtain useful $\MTL$ formulas, we identify a number of 
MTL patterns, listed in Table~\ref{tab:MTL-patterns}, that are commonly used for monitoring cyber-physical systems.
For instance, the time-sensitive requirement of an electronically controlled steering (ECS) system ``operational checks like RAM verification must be done every 20 secs" can be monitored globally using the bounded recurrence formula $\lF_{[0,20]} \mathtt{operational\_check}$~\cite{mtlformulas}; the requirement of an autonomous vehicle (from the introductory example) ``brake should be triggered until within 2 secs the vehicle has no obstacle in an unsafe distance " can be monitored globally using the bounded until formula $\mathtt{brake} \lU_{[0,2]} \mathtt{no\_obstacle}$.
% The patterns bounded recurrence, bounded response, and bounded invariance are used to describe time-sensitive requirements of an electronically controlled steering (ECS) system.
% The 
% We included an additional $\MTL$ pattern consisting of the $\lU$-operator to showcase the performance of \tool{} on formulas with $\lU$. 

% \begin{table}[!ht]
%     \vspace{-7mm}
% 	\centering
% 	\caption{$\MTL$ requirements for Autonomous Vehicle System}
% 	\label{tab:MTL-patterns}
% 	%\resizebox{\linewidth}{!}{%
% 	\begin{tabular}{|l|}
% 		\hline
% 		\rule{0pt}{1.0\normalbaselineskip}  
%             $\lG(\neg s \rightarrow  \lF_{[0,2]} b)$ \\[1mm]
%             $\lG(\neg s \rightarrow (b\lU_{[0,2]} s))$ \\[1mm]
%             $\lG(\lG_{[0,2]} s \rightarrow \lF_{[2,3]}\neg b)$ \\[1mm]
% 		\hline
% 	\end{tabular}
% \end{table}

\vspace{-6mm}
\begin{table}[!ht]
	\centering
	\caption{Typical MTL patterns used for monitoring cyber-physical systems}
	\label{tab:MTL-patterns}
        \vspace{2mm}
	%\resizebox{\linewidth}{!}{%
	\begin{tabular}{|rl|}
		\hline
		\rule{0pt}{1.0\normalbaselineskip} Bounded Recurrence: & \textit{Globally}$(\lF_{[t_1,t_2]}p)$ \\[1mm]
		Bounded Response: & \textit{Globally}$(p \rightarrow \lF_{[t_1,t_2]}q)$ \\[1mm]
		Bounded Invariance: & \textit{Globally}$(p \rightarrow \lG_{[t_1,t_2]}q)$ \\[1mm]
        Bounded Until: & \textit{Globally}$(p \lU_{[t_1,t_2]} q)$ \\[1mm]
		\hline
	\end{tabular}
\end{table}
\vspace{-3mm}

In our experiments, we construct $\MTL$ formulas from the patterns in Table~\ref{tab:MTL-patterns} by replacing time interval $[t_1,t_2]$ with different values.
Now, to generate a sample from an $\MTL$ formula $\varphi$, we generated a set of random prefixes and then classified them into positive or negative depending on whether $\varphi$ holds at all time-points of the prefix or not.
We conducted all the experiments on a single core of a AMD EPYC 7702 64-Core CPU (at 2GHz) using up to 10GB of RAM.
The timeout was set to be 5400 secs for all the experiments.

To address \textbf{RQ1}, we ran \tool{} on a benchmark suite generated from nine $\MTL$ formulas obtained from the three $\MTL$ patterns in Table~\ref{tab:MTL-patterns} by replacing $t_1$ with 0 and $t_2$ with 1,2, and 3.
The suite consists of 36 samples for each pattern (12 samples for each formula), with the number of prefixes ranging from 10 to 40 and the length of prefixes (i.e., the number of sampled time points) ranging from 4 to 6.
For each sample $\sample$, we set the future-reach bound $\bound$ to be $\fr(\varphi)$, where $\varphi$ is the formula from which $\sample$ was generated. 
\vspace{-6mm}
\begin{table}[h]
\centering
\caption{Summary of the synthesized formulas.}
\label{tab:recovery}
\setlength{\tabcolsep}{4pt}
\begin{tabular}{cccccc}
\toprule
Formula pattern & \multicolumn{2}{c}{Successful runs} & Timed out & Avg Size & Avg Time \\
\cmidrule{2-3}
 & Matched & Not Matched & & & (in sec) \\
\midrule
Bounded Recurrence  & 36 & 0 & 0 & 2 & 17.5 \\
Bounded Response  & 25 & 5 & 6 & 3.7 & 1860.3 \\
Bounded Invariance  & 15 & 7 & 14 & 3.6 & 1397.2 \\
Bounded Until & 32 & 4 & 0 &  2.9 & 362.4 \\
\bottomrule
\end{tabular}
\end{table}

We depict the summary of the results for this experiment in Table~\ref{tab:recovery}.
For each run, we noted the formula synthesized, its size and the total time taken.
Further, we noted whether the synthesized formula matched the pattern of the original formula using which the sample was generated.
We observed that the synthesized formulas matched the pattern of the original formula in 87.1\% of the cases in which \tool{} did not time out.
This shows that the randomly generated samples captured the behaviour of the original formula rather well, enabling a fair evaluation of \tool{}.

Furthermore, we observed that the size of the synthesized
formula is always equal to or less than that of the original formula, demonstrating that \tool{} always finds a concise formula for a given future-reach bound
Thus, we answer RQ1 in positive.

%we noted the number of cases in which \tool{} recovered a formula of the original $\MTL$ pattern and the number of cases in which it timed out.
%From the table, we observe that among the cases in which \tool{} did not time out, it returned a formula of the original pattern in  of the cases.

To address \textbf{RQ2}, we investigate how the size of the synthesized formula changed over varying future-reach bounds.
For this, we ran \tool{} on the same benchmark suite from RQ1 but, this time, by varying the future-reach bound $\bound$ from 1 to 4. 
We investigate the average size of the minimal formula we get over the generated 108 samples for each future-reach bound. 
% \begin{table}
%     \centering
%     \caption{Average size of the smallest formula over 120 samples for varying future-reach bound}
% 	\label{tab:size_fr}
%     \vspace{0.1cm}
% 	\begin{tabular}{|c@{\hskip 1em}|c@{\hskip 1em} |c@{\hskip 1em}|c@{\hskip 1em}|c@{\hskip 1em}|}
% 		\hline
%             Future-Reach bound ($k$) &1 &2 &3 &4\\
%             \hline
% 		Average Size of formula  & 3.614  &3.461 & 3.133 &3.125 \\
% 		\hline
% 	\end{tabular}
% \end{table}

We observed that for future-reach bounds $\bound$ of 1, 2, 3, and 4, the average size of the synthesized minimal formulas were 3.904, 3.734, 3.370, and 3.361, respectively.
Thus, the trend is that with an increase in $\bound$, the average size of the minimal formula decreased. 
This is because an increase in $\bound$ allows a bigger search space of formulas.
One can, however, also notice that the decrease in the average size of the formulas with increasing future-reach bound is not vast.
This highlights the advantage of using a future-reach bound for synthesizing formulas for online monitoring and confirms the efficacy of our algorithm.

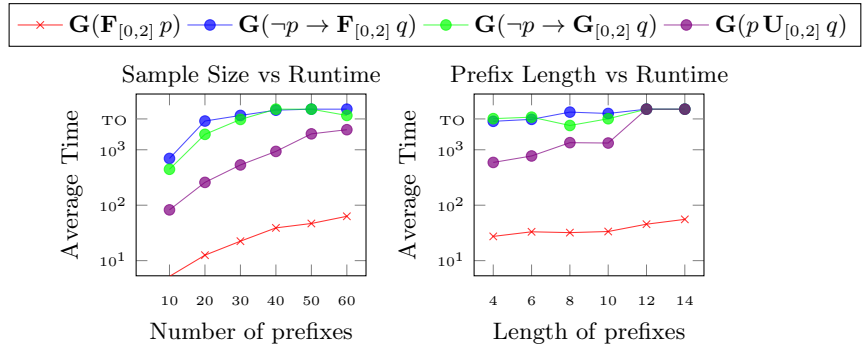
\begin{figure}[t]
\begin{center}
{
\begin{subfigure}[b]{0.4\textwidth}
        \begin{tikzpicture}
            \begin{semilogyaxis}[%[log origin=infty] 
            height=40mm,
            xmin= 6, ymin=10,
            enlarge x limits=true, enlarge y limits=true,
            xlabel = {Number of prefixes},
            ylabel = {Average Time},
            xtick={10,20,30,40,50,60},%
            ytickten= {0, 1, 2, 3},%
            %extra x ticks={2000}, extra x tick labels={\strut TO},%
            extra y ticks={3600}, %extra y tick labels=\empty%, 
            extra y tick labels={\strut TO},
            xminorticks=false,
            yminorticks=false,
            xlabel near ticks,
            ylabel near ticks,
            %xtick style={draw=major ticks},
            %ytick style={draw=none},
            label style={font=\footnotesize},
            x tick label style={font={\strut\tiny}},
            y tick label style={font={\strut\tiny}},		
            x label style = {yshift=1mm},
            legend columns = 4,
            legend style={at={(1.3,1.5)},
                anchor=north,fill=none},
            title = {Sample Size vs Runtime},
            title style = {font=\footnotesize, inner sep=-4pt}
            ]
              
            \addplot[color=red,mark=x,opacity =0.8]
            plot coordinates {
                (10,5.20)
                (20,12.55)
                (30,22.44)
                (40,38.99)
                (50,46.8)
                (60,63.10)
            };
             \addlegendentry{$\lG(\lF_{[0,2]}p)$}

            \addplot[mark=*,blue,opacity =0.7] plot coordinates {
                (10,697.23)
                (20,3297.45)
                (30,4159.03)
                (40,5176.83)
                (50,5400)
                (60,5400)
            };
            \addlegendentry{$\lG(\neg p \rightarrow \lF_{[0,2]}q)$}

             \addplot[mark=*,green,opacity =0.7] plot coordinates {
                (10,444.51)
                (20,1897.77)
                (30,3524.84)
                (40,5400)
                (50,5400)
                (60,4158.06)
            };
            \addlegendentry{$\lG(\neg p \rightarrow \lG_{[0,2]}q)$}

            \addplot[mark=*,violet,opacity =0.7] plot coordinates {
                (10,82.33)
                (20,257.97)
                (30,532.59)
                (40,933.82)
                (50,1936.27)
                (60,2300.62)
            };
            \addlegendentry{$\lG(p \lU_{[0,2]} q)$}
            
            \end{semilogyaxis}
            \end{tikzpicture}
            \end{subfigure}
            \hskip 1mm
            \begin{subfigure}[b]{0.4\textwidth}
        \begin{tikzpicture}
            \begin{semilogyaxis}[%[log origin=infty] 
            height=40mm,
            xmin= 4, ymin=10,
            enlarge x limits=true, enlarge y limits=true,
            xlabel = {Length of prefixes},
            ylabel = {Average Time},
            xtick={4, 6, 8, 10, 12, 14},%
            ytickten= {0, 1, 2, 3},%
            %extra x ticks={2000}, extra x tick labels={\strut TO},%
            extra y ticks={3600}, %extra y tick labels=\empty%, 
            extra y tick labels={\strut TO},
            xminorticks=false,
            yminorticks=false,
            xlabel near ticks,
            ylabel near ticks,
            %xtick style={draw=major ticks},
            %ytick style={draw=none},
            label style={font=\footnotesize},
            x tick label style={font={\strut\tiny}},
            y tick label style={font={\strut\tiny}},		
            x label style = {yshift=1mm},
            legend columns = 6,
            legend style={at={(0,1.6)},
                anchor=north,fill=none},
            title = {Prefix Length vs Runtime},
            title style = {font=\footnotesize, inner sep=-4pt}
            ]
              
            \addplot[color=red,mark=x,opacity =0.8]
            plot coordinates {
                (4.0,27.371908)
                (6.0,33.134485)
                (8.0,32.022461)
                (10.0,33.536167)
                (12.0, 45.45)
                (14.0, 55.55)
                
            };
             % \addlegendentry{$\lG(\lF_{[0,2]}p)$}
         % 
            
            \addplot[mark=*,blue,opacity =0.7] plot coordinates {
                (4.0,3260.98)
                (6.0,3536.31)
                (8.0,4776.90)
                (10.0,4513.69)
                (12.0, 5400)
                (14.0, 5400)

                };
            % \addlegendentry{$\lG(p \lor \lF_{[0,2]}q)$}
        \addplot[mark=*,green,opacity =0.7] plot coordinates {
                (4.0,3642.89)
                (6.0,3875.82)
                (8.0,2739.82)
                (10.0,3624.90)
                (12.0, 5400)
                (14.0, 5400)
                };

        \addplot[mark=*,violet,opacity =0.7] plot coordinates {
                (4.0,590.64)
                (6.0,772.78)
                (8.0,1343.12)
                (10.0,1322.52)
                (12.0, 5400)
                (14.0, 5400)
                };
            % \addlegendentry{$\lG(p \lor \lG_{[0,2]}q)$}            
            \end{semilogyaxis}
            \end{tikzpicture}
            \end{subfigure}
           }
        \end{center}
        \vspace{-5mm}
	\caption{Runtime change with respect to the number of prefixes and prefix lengths}
	\label{fig:scale_results}
\end{figure}

To address \textbf{RQ3}, we ran \tool{} on a benchmark suite generated from $\MTL$ formulas which originate from the $\MTL$ patterns in Table~\ref{tab:MTL-patterns}, setting $t_1=0$ and $t_1=2$. 
The suite consists of 36 samples for each formula, with the number of prefixes varying from 10 to 60 and the length of prefixes varying from 4 to 14.
We set the future-reach bound $\bound$ to be two. 
%The number of signal prefixes in our benchmark suite ranges from \red{nr.} to \red{nr.}, and the length of the signal prefixes range from \red{nr.} to \red{nr.}.

Figure~\ref{fig:scale_results} illustrates the runtime variation of \tool{} in two cases: increasing the number of prefixes fixing the length of them and increasing the length of prefixes fixing the number of them.
We observe that to synthesis a larger formula the time required grows significantly.
This trend can be noticed in both the figures.

%We observe that the increase in runtime is steeper relative to the prefix lengths than to the number of prefixes. 
%This is natural because an increase in prefix length increases $\mu$, which significantly affects the size of the encoding. 
%We notice an anomalous drop in runtime for the third formula for prefix length 12.
%A plausible explanation is that the randomly generated samples with prefixes of length beyond 12 could not capture the property of the formula properly and allowed simpler formulas.

%% file: conclusion.tex
We have presented a novel SMT-based algorithm for automatically synthesizing $\MTL$ specifications from finite system executions. 
To be useful for efficient monitoring, we ensure that the synthesized formulas are both concise and have low future-reach.
We have shown that our algorithm can synthesize concise formulas from benchmarks generated from commonly used $\MTL$ patterns.

While our algorithm is tailored to synthesize globally separating formulas particularly useful for monitoring, we can adapt our algorithm easily to synthesize only \emph{separating} formulas as in the standard temporal logic inference 
 setting~\cite{flie,MohammadinejadD20}.
Our algorithm includes all the standard temporal operators that are typically used in $\MTL$.
However, we believe it is possible to improve the performance of the algorithm by omitting a temporal operator such as $\lU_I$ for which the encoding can be substantially large.

From a practical point of view, an interesting future direction will be to lift our techniques to automatically synthesize $\STL$ formulas for verification. A straightforward approach towards this using the above-mentioned constraint-based methods has been explained in~\cite{Raha23}. However, for industrial use and scalability, clever heuristics and optimizations are needed to be explored in future work.

%% file: appendix.tex
\label{sec:appendix}